\documentclass[preprint,3p,12pt,authoryear]{elsarticle}
\usepackage{color}
\usepackage{amssymb, amsfonts, amsthm, amsmath}
\usepackage{graphicx, epstopdf}
\usepackage{cases}
\usepackage{hyperref}
\usepackage{cleveref}
\usepackage{thmtools} 
\usepackage{tikz}
\usepackage[utf8]{inputenc}
\usepackage{bbm}
\usepackage{caption}
\usepackage{subcaption}
\usepackage{bm}
\usepackage{empheq}
\usepackage{multirow}
\usepackage{mathtools}
\usepackage{natbib}
\usepackage{booktabs}
\usepackage{tabularx}

\theoremstyle{plain}
\newtheorem{theorem}{Theorem}[section]
\newtheorem{corollary}[theorem]{Corollary}

\newtheorem{lemma}[theorem]{Lemma}
\newtheorem{assumption}[theorem]{Assumption}
\newtheorem{definition}[theorem]{Definition}
\newtheorem{remark}[theorem]{Remark}
\newtheorem{example}[theorem]{Example}

\numberwithin{equation}{section}
\makeatletter
\@ifundefined{theHequation}
  {}
  {}
\@ifundefined{theHparentequation}
  {}
  {}
\AtBeginEnvironment{subequations}{%
  }
\AtEndEnvironment{subequations}{%
  }
\@ifundefined{theHtheorem}
  {}
  {}
\@ifundefined{theHcorollary}
  {}
  {}
\@ifundefined{theHproposition}
  {}
  {}
\@ifundefined{theHlemma}
  {}
  {}
\@ifundefined{theHassumption}
  {}
  {}
\@ifundefined{theHdefinition}
  {}
  {}
\@ifundefined{theHremark}
  {}
  {}
\@ifundefined{theHexample}
  {}
  {}
\makeatother

\usepackage{algorithm}
\usepackage{algpseudocode}

\makeatletter
\DeclareRobustCommand{\borderlesshref}[2]{%
  \begingroup
  \def\@pdfborder{0 0 0}%
  \href{#1}{#2}%
  \endgroup}
\makeatother
\journal{}
\begin{document}
\begin{frontmatter}

    \title{Nonlinear Equilibrium Transitions in a Potential Game Model for Federated Learning}
    \nonumnote{Published in \textit{Physica D: Nonlinear Phenomena}, Volume 495 (October 2026), Article 135288. DOI: \href{https://doi.org/10.1016/j.physd.2026.135288}{10.1016/j.physd.2026.135288}.}
	
	\author[label1]{Kang Liu}
	\ead{kang.liu@u-bourgogne.fr}
	\author[label2]{\borderlesshref{https://iziqi.github.io/}{Ziqi Wang}\corref{cor}}
	\cortext[cor]{Corresponding author. Authors are listed alphabetically by family name.}
	\ead{ziqi.wang@fau.de}
	\author[label2,label3,label4]{Enrique Zuazua}
	\ead{enrique.zuazua@fau.de}

        \affiliation[label1]{
		organization={Institut de Mathématiques de Bourgogne, Université Bourgogne Europe, CNRS},
		city={Dijon},
		postcode={21000}, 
		country={France}}

	\affiliation[label2]{
		organization={Chair for Dynamics, Control, Machine Learning and Numerics – Alexander von Humboldt Professorship, Department of Mathematics, Friedrich-Alexander-Universität Erlangen-Nürnberg},
		addressline={Cauerstrasse 11}, 
		city={Erlangen},
		postcode={91058}, 
		country={Germany}}
	
	\affiliation[label3]{
		organization={Chair of Computational Mathematics, Fundación Deusto},
		addressline={Avenida de las Universidades, 24}, 
		city={Bilbao},
		postcode={48007}, 
		state={Basque Country},
		country={Spain}}
	
	\affiliation[label4]{
		organization={Departamento de Matemáticas, Universidad Autónoma de Madrid},
		addressline={Ciudad Universitaria de Cantoblanco}, 
		city={Madrid},
		postcode={28049}, 
		country={Spain}}
        
    \begin{abstract}
    In federated learning (FL), a central server typically allocates training efforts to clients. However, from a market-oriented perspective, clients may independently choose their training efforts based on rational self-interest. To study this setting, we propose a potential game framework in which each client's payoff is determined by its individual effort and the rewards provided by the server. The rewards are influenced by the collective efforts of all clients and can be modulated by a reward factor. We first establish the existence of Nash equilibria (NEs) and then investigate their uniqueness in a stationary setting. We show that the NEs depend nonlinearly on the reward factor and exhibit a nonsmooth transition at a critical value, where the stationary potential loses strict curvature, leading to nonunique NEs and a jump between low-effort and high-effort branches. Furthermore, we prove the convergence of the best-response algorithm for computing NEs in our FL game. Finally, we apply the clients' rational efforts derived from the NEs to FL training with various datasets and models, thereby validating the effectiveness of the identified critical reward factor.
    \end{abstract}
	
	
	
	
\end{frontmatter}

\section{Introduction}\label{sec_intro}
Federated learning (FL) \citep{mcmahan2017communicationefficient} has shown considerable promise for training large-scale machine learning models across distributed data sources. 
In FL, clients train local models using local datasets and transmit only the updated model parameters, rather than the raw data, to a central server for aggregation. 
{From the viewpoint of dynamical systems, the iterative client-server updates can be viewed as a discrete approximation of a coupled gradient flow, where each client's local descent contributes to the evolution of the global model parameters along a shared loss landscape.}

\subsection{Motivations}\label{subsec:motivation}

\begin{figure}[h]
\centering
\includegraphics[width=0.75\textwidth]{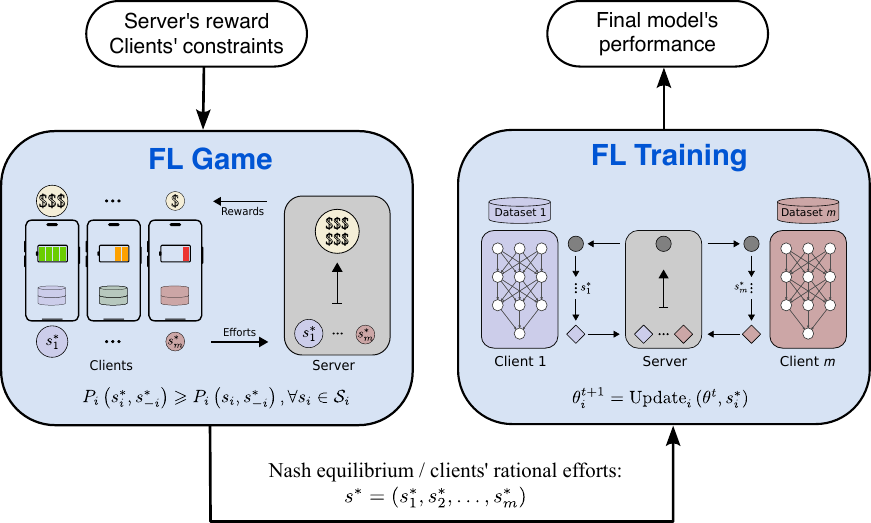}
\caption{In the FL game (left), each client $i$ maximizes the payoff $P_i$, and the NE $s^*$ satisfies $P_i(s_i^*, s_{-i}^*) \ge P_i(s_i, s_{-i}^*)$ for all $s_i \in \mathcal{S}_i$. The equilibrium efforts $s^*$ (\textit{e.g.}, local epochs) then drive the FL training stage (right), where the model parameters are updated according to $\theta_i^{t+1} = \operatorname{Update}_i(\theta^t, s_i^*)$.}
\label{fig_FL_game_train}
\end{figure}

\paragraph{Incentives as a game-theoretic interaction} Conventional FL often assumes voluntary participation from heterogeneous clients while the server prescribes local workloads in a {centrally planned} manner.
When private costs differ across clients, this can induce strategic under-training or silent dropout \citep{hard2018federated}.
We therefore model incentives as a game-theoretic interaction. The server offers rewards, each client chooses its effort by balancing the reward against its local cost, and the coupled FL aggregation makes each payoff depend on the efforts of others.
This yields a Nash equilibrium (NE) characterization of a self-consistent effort profile, and \Cref{fig_FL_game_train} illustrates how this equilibrium is integrated into the FL training loop.
Guided by this viewpoint, we study two sets of questions:
\begin{itemize}
\item How should the FL game be modeled (reward rule, local costs, and feasible sets), and under what assumptions do NE existence, uniqueness, and efficient computation hold?
\item How does the equilibrium effort profile respond to the reward parameter? In particular, do threshold-type or nonlinear transitions arise, and how should the server tune this reward parameter to improve participation quality and final model performance?
\end{itemize}

\paragraph{{Rationale for potential games}}
Analyzing NEs in general $N$-player games is notoriously challenging \citep{daskalakis2009complexity}, and tracing how equilibria vary with game parameters is even harder \citep{hofbauer2009stable}.
In FL, clients' updates are typically aggregated through weighted sums, which motivates designing the server's reward as a compatible weighted sum of client efforts.
Under this structure, the interaction becomes a \emph{potential game} \citep{monderer1996potential}, allowing equilibria to be studied through a single potential function and analyzed with standard optimization tools. 

\subsection{Main Results}
\begin{figure}[ht]
\centering
\includegraphics[width=0.47\textwidth]{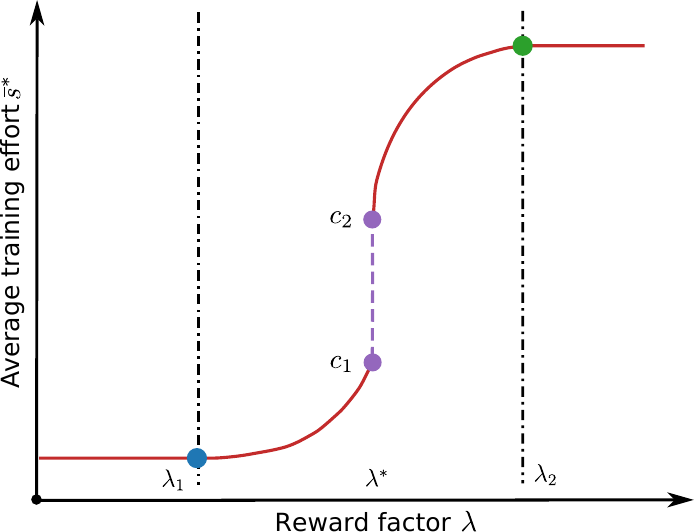}
\caption{Illustration of the nonlinear threshold structure and jump phenomenon.}
\label{fig_thresholds}
\end{figure}

We state the core findings first and defer the full model setup to Section~\ref{sec_FL_potential_game}.
\begin{enumerate}
\item \textbf{Existence via potential structure.} We formulate client participation in FL as a weighted potential game and prove the existence of NEs in Theorem~\ref{thm_exist}.

\item \textbf{Uniqueness in the stationary quadratic regime.} In Theorem~\ref{thm_uniq}, for stationary efforts and quadratic costs, we show uniqueness for all values of the server's reward factor $\lambda\neq\lambda^*$, while at $\lambda=\lambda^*$ the game may admit infinitely many equilibria.

\item \textbf{Nonlinear structure and landscape transition.} Corollary~\ref{cor_uniq} shows that the average training effort $\bar{s}^*$ grows nonlinearly with respect to $\lambda$, with three thresholds $\lambda_1<\lambda^*<\lambda_2$ for activation, jump, and saturation (see \Cref{fig_thresholds} and \Cref{tab_critical_constants}). In particular, at $\lambda=\lambda^*$, the curve undergoes a nonsmooth transition. \Cref{rem_geom_lambda_star} explains this by the loss of strict curvature and the emergence of a flat direction in the stationary potential, leading to nonunique NEs and a jump in the equilibrium effort.

\item \textbf{Best-response computation.} In Theorem~\ref{thm_BRA} (and the general Theorem~\ref{thm_BRA_general}), we establish convergence guarantees for the best-response algorithm used to compute NEs.

\item \textbf{Theory-experiment consistency.} Across multiple datasets and FL methods, the strongest performance gain occurs near the jump regime around $\lambda^*$, consistent with the nonlinear sensitivity predicted by the theory.

\end{enumerate}

\vspace{-18pt}

\begin{table}[h]
\centering
\caption{Summary of the critical thresholds, see Theorem~\ref{thm_uniq} and \Cref{cor_uniq} for more details.}
\label{tab_critical_constants}
\small
\begin{tabularx}{\textwidth}{c l c >{\raggedright\arraybackslash}X}
\toprule
Symbol & Name & Eq. & Definition and role \\
\midrule
$\lambda^*$ & Jump point & \eqref{eq_lambda_star} & The critical reward factor where non-uniqueness occurs. \\
$\lambda_1$ & Activation point & \eqref{eq_lambda_1_2} & If $\lambda \in (0,\lambda_1)$, the unique NE is at minimum efforts. \\
$\lambda_2$ & Saturation point & \eqref{eq_lambda_1_2} & If $\lambda \in (\lambda_2, +\infty)$, the unique NE is at maximum efforts. \\
$c_1$ & Lower jump bound & \eqref{constant_C1_C2} & If $\lambda \in (\lambda_1,\lambda^*)$, the unique NE satisfies $\bar{s}^*<c_1$. \\
$c_2$ & Upper jump bound & \eqref{constant_C1_C2} & If $\lambda \in (\lambda^*,\lambda_2)$, the unique NE satisfies $\bar{s}^*>c_2$. \\
\bottomrule
\end{tabularx}
\end{table}

\subsection{Related Work}\label{sec_related_work}

Game-theoretic analysis has long been used for multi-agent decision-making \citep{barron2013game, osborne1994course}.
In FL, it is widely used to study fairness, robustness, and strategic participation \citep{gupta2023federated}.
Existing FL game formulations emphasize different objects (coalitions, payments, or contribution scores), which leads to different analytical guarantees.
Below we position our formulation against representative lines of work.

\subsubsection{Coalitional Game Theory for Fairness in FL}
Coalitional (hedonic) games \citep{banerjee2001core,bogomolnaia2002stability} model how clients form stable groups.
In FL, model-sharing formulations \citep{donahue2021model,donahue2021optimality} analyze stability and welfare gaps between stable and socially optimal coalitions.
These models are well suited to fairness questions driven by grouping and coalition stability, where the key variable is the coalition partition itself.
These works focus on a discrete coalition structure rather than a continuous equilibrium effort profile controlled by a server-side reward parameter. As a result, they do not provide our type of NE uniqueness and best-response convergence results.

\subsubsection{Stackelberg and Auction-Based Incentive Mechanisms}\label{rw:stackelberg}
Stackelberg and auction-based mechanisms design payments to induce participation from strategic clients.
Representative works study bi-level reward allocation \citep{khan2020federated}, closed-form market equilibria and multi-tier extensions \citep{lee2020market,dong2020federated}, and practical marketplace systems with security, reputation, or privacy guarantees \citep{li2023martfl,li2024flauction,lu2024privdata}.
Their main strength is explicit economic mechanism design at the server level, often with implementable pricing or allocation rules.
These studies mainly characterize payment/allocation rules, whereas our focus is a full threshold analysis of equilibrium effort profiles under a server-tunable reward parameter.

\subsubsection{Shapley Value-based Contribution Evaluation}\label{rw:valuation}
Shapley-based methods evaluate participant contributions after training \citep{shapley1953value}.
Recent work improves scalability via gradient replay and sampling approximations \citep{song2019profit,liu2022gtg,yu2020fairness}.
This line is particularly useful for retrospective attribution and reward sharing once model updates are already observed.
Because they are post-training valuation methods, they do not analyze strategic effort selection before training, equilibrium thresholds, or best-response computation for NE profiles.

\subsubsection{{Potential Games and Their Applications}}
Potential games \citep{monderer1996potential} offer a useful structure because NEs correspond to stationary points of one potential function, which often enables tractable best-response analysis \citep{swenson2018best}.
This framework has been effective in areas such as multi-agent control and resource allocation \citep{liu2019game,britzelmeier2019numerical,wu2020potential,raschella2020dynamic}.
For incentive problems, this structure is especially attractive because equilibrium existence and dynamics can be studied through one global scalar function.
Compared with the FL incentive formulations reviewed above, our contribution is formulated within a single server-tunable game. Within this framework, we establish equilibrium uniqueness, characterize a nonlinear reward threshold structure, and show convergence of best-response dynamics. To the best of our knowledge, these three properties have not been established jointly in the FL incentive formulations reviewed above.

\subsection{Notation}
In this subsection, we introduce the key notation used in the remainder of the paper.
We denote by $\Gamma$ a game with three components. The first component is a finite set of players $[m] \coloneqq \{1, \ldots, m\}$. The second is a pure strategy set $\mathcal{S}_i \subseteq \mathbb{R}^d$ for each player $i\in[m]$, together with $\mathcal{S} \coloneqq \prod_{i\in [m]} \mathcal{S}_i$. The third is a payoff function $P_i \colon \mathcal{S} \to \mathbb{R}$ for each player $i\in[m]$.
We say that a game $\Gamma$ is finite if its strategy set $\mathcal{S}$ is finite. 
For any $i\in [m]$, we define $\mathcal{S}_{-i} \coloneqq \prod_{j\neq i} \mathcal{S}_j$.
For convenience, for any $s \in \mathcal{S}$, we do not distinguish between $P_i(s)$ and $P_i(s_i, s_{-i})$, where
$s_i\in \mathcal{S}_i$ is the $i$-th coordinate of $s$ and $s_{-i} \in \mathcal{S}_{-i}$ is the tuple of other coordinates.

In a non-cooperative game, each player $i \in [m]$ is self-interested and seeks to maximize their payoff $P_i(s_i, s_{-i})$, leading to the following concept of an NE.

\begin{definition}[Nash equilibrium]
In the game $\Gamma$, we call a point $s^{*}\in \mathcal{S}$ an NE if the following inequality holds:
\begin{equation*}
    P_i\left(s_i^{*}, s_{-i}^{*}\right) \geqslant P_i\left(s_i, s_{-i}^*\right), \quad  \forall s_i \in \mathcal{S}_i, \, \forall i \in [m].
\end{equation*}
\end{definition}

\noindent A special type of non-cooperative games is known as potential games \citep{monderer1996potential}, which can be defined as follows:
{\begin{definition}[Potential game]\label{def_potential_game}
Let $ w = (w_i)_{i \in [m]} $ be a strictly positive vector.
The game $\Gamma$ is said to be a (weighted) potential game if there exists a function $ P \colon \mathcal{S} \to \mathbb{R} $ such that
\begin{equation*}
    P_i(s_i, s_{-i}) - P_i(x_i, s_{-i}) = w_i \left( P(s_i, s_{-i}) - P(x_i, s_{-i}) \right), \quad \forall s_i, x_i \in \mathcal{S}_i, \, \forall i \in [m].
\end{equation*}
\end{definition}}

\subsection{Outline of the Paper}
The remainder of the paper is organized as follows. In \Cref{sec_FL_potential_game}, we present the design of our FL game and establish the existence of an NE\@. Next, in \Cref{sec_homo_game_uniq}, we focus on the uniqueness of the NE in a stationary scenario and study the convergence of the best-response algorithm. Simulation results are presented and analyzed in \Cref{sec_numrical}, followed by detailed technical proofs and theorems in \Cref{sec_proof}. Finally, we summarize our conclusions and perspectives in \Cref{sec_conclusions}.

\section{A Potential Game Framework for FL}\label{sec_FL_potential_game}

This section models client effort selection in FL as a potential game. We first recall the FL updates, then define the FL game, and finally establish the existence of NEs through the potential structure.

We start from the standard FL training objective:
\begin{equation}\label{prob_FL}
\min_{\theta \in \mathbb{R}^n} \sum_{i=1}^m \rho_i \ell_i(\theta), \text{ where } \rho_i = \frac{D_i}{D}, \text{ for } i \in [m],
\end{equation}
where $m$ is the number of clients, $D_i$ is client $i$'s local data size, $D=\sum_{i=1}^m D_i$, and $\ell_i$ is the local empirical loss.
FL training proceeds over communication rounds with local updates and server aggregation. At round $t \in [T]$, each client $i \in [m]$ initializes $\theta_i^{t,0}=\theta^t$ and performs
\begin{equation}\label{eq_gradient_descent}
\theta_i^{t,e}=\theta_i^{t,e-1}-\eta_i\nabla\ell_i(\theta_i^{t,e-1}),\quad e=1,\ldots,s_i^t,
\end{equation}
then sends $\theta_i^{t+1}\coloneqq\theta_i^{t,s_i^t}$ to the server. The server aggregates
$\theta^{t+1}=\sum_{i=1}^m \rho_i \theta_i^{t+1}$ and broadcasts $\theta^{t+1}$ for the next round.

Rather than assuming that the server prescribes client efforts $s_i^t$, we model clients as self-interested players who choose effort by balancing rewards against costs.
Accordingly, we adopt a finite-player game framework to characterize clients' rational effort choices.
In this game, each client maximizes a payoff determined by local training cost and server reward.
We denote this game as $\Gamma_{\textnormal{FL}}$ and provide its definition below.
\begin{definition}[FL game $\Gamma_{\textnormal{FL}}$]
\label{def_FL_game}
The FL game $\Gamma_{\textnormal{FL}}$ is specified by the player set
$[m]$, the strategy profile
$\mathcal{S}=\prod_{i\in[m]}\mathcal{S}_i$ with
$\mathcal{S}_i\subseteq\mathbb{R}_+^T$, and payoff functions
$P_i\colon\mathcal{S}\to\mathbb{R}$ of the form
\begin{equation}\label{eq_payoff_P_i_intro}
    P_i(s_i, s_{-i}) = \sum_{t=1}^T r_{\lambda}(s_i^t,s_{-i}^t) - c_i(s_i^t), \quad i\in[m],
\end{equation}
where $r_{\lambda}(s_i^t,s_{-i}^t)$ is the server-side reward, and $c_i(s_i^t)$ denotes client $i$'s local cost.
\end{definition}

As shown in \eqref{eq_payoff_P_i_intro}, by adjusting $\lambda$, the server can steer the equilibrium profile $s^*(\lambda)$.
Next, we introduce the three core components of the FL game, namely strategy sets in \Cref{subsec_strategy_set}, reward design in \Cref{subsec_game_design}, and cost design in \Cref{subsec_cost_c_i}.

\subsection{Clients' Strategy Sets}
\label{subsec_strategy_set}

We measure each client's effort by the number of local update steps (or, equivalently, epochs under a fixed conversion).
In practical FL systems, client efforts are bounded by resource constraints, and we consider the following two feasible-set models.

\begin{example}[Total budget case]\label{eg_hetro}
For client $i$, let
\begin{equation}\label{eq_hetro_S}
\mathcal{S}_i = \left\{ s_i \in \mathbb{R}_{+}^T \, \Big|\, b_i\leq \sum_{t\in [T]} s_i^t \leq B_i\right\},
\end{equation}
where $b_i$ and $B_i$ are the minimum and maximum total efforts over $T$ rounds.
\end{example}

\begin{example}[Stationary case]\label{eg_homo}
In many FL scenarios, clients select one effort level and keep it fixed over rounds. Then
\begin{equation}\label{eq:homo_S}
\mathcal{S}_i = \left\{ s_i \in \mathbb{R}_{+}^T \, \Big|\, q_i\leq s_i^t = s_i^{\tau}\leq Q_i, \, \forall \,t, \tau \in [T] \right\},
\end{equation}
where $q_i$ and $Q_i$ are the minimum and maximum per-round efforts.
\end{example}

\subsection{Design of the Reward $r_{\lambda}(s^t_i, s^t_{-i})$}
\label{subsec_game_design}\label{subsec_reward_r_i}

We consider the incentive rule
\begin{equation}
\label{eq_reward1}
    r_{\lambda}(s_i^t,s_{-i}^t) =  p_{\lambda}^t s_i^t, \quad \text{where }\, p_{\lambda}^t = \lambda  \sum_{j=1}^m \rho_j s_j^t.
\end{equation}
Here, $\lambda>0$ is the server-controlled reward factor.
Since the unit price $p_{\lambda}^t$ increases with clients' aggregate effort, each client's marginal incentive is strengthened by the efforts of others, creating strategic complementarity among clients. This is consistent with FL evidence that greater effective local computation can improve performance under fixed communication budgets \citep{kairouz2021advances,li2020federateda, mcmahan2017communicationefficient}.

\subsection{Design of the Cost $c_i(s_i^t)$}
\label{subsec_cost_c_i}

The function $c_i(s_i^t)$ quantifies client $i$'s participation burden.
A linear model $c_i(s_i^t)=\alpha_i s_i^t$ represents constant marginal cost.
However, if we interpret the cost as a measure of the client's reluctance to exert effort, a quadratic function, $c_i(s_i^t) = \alpha_i (s_i^t)^2$, is more appropriate. This captures the tendency of clients to become increasingly unwilling to exert higher levels of local training effort. For instance, on a smartphone, performing the first few local updates when the battery is full may be perceived as inexpensive, whereas executing the same number of updates under low battery appears significantly more costly. Such nonlinear perceptions of cost are well captured by the quadratic model.

\subsection{Existence of an NE of $\Gamma_{\text{FL}}$}\label{subsec_modeling}

Based on the formulation of the strategy set and payoff functions, we now present the results regarding the potential structure and the existence of an NE in the game $\Gamma_{\text{FL}}$.

\begin{theorem}[Existence]\label{thm_exist}
Consider the game $ \Gamma_{\textnormal{FL}}$ (see \Cref{def_FL_game}) with the reward function \eqref{eq_reward1}. Then it admits a $w$-potential given by
\begin{equation}\label{potential_func}
    P_{\textnormal{FL}}(s) = \sum_{i=1}^m \sum_{t=1}^T \left(\frac{\lambda}{2}\rho_i^2(s_i^t)^2 - \rho_i c_i(s_i^t)\right) + \sum_{t=1}^T \frac{\lambda}{2}\left(\sum_{i=1}^m \rho_i s_i^t\right)^2, \quad s \in \mathcal{S},
\end{equation}
where $c_i(\cdot)$ is client $i$'s local cost, and $w_i = 1/\rho_i$ for $i \in [m]$.
If $\mathcal{S}$ is compact and $c_i(\cdot)$ is lower semi-continuous, then $ \Gamma_{\textnormal{FL}} $ possesses at least one NE.
\end{theorem}
\begin{proof}
It is straightforward to verify that $P_{\textnormal{FL}}$ given in \eqref{potential_func} is a $w$-potential of $\Gamma_{\textnormal{FL}}$ in the sense of \Cref{def_potential_game}, with $w_i = 1/\rho_i$ for all $i \in [m]$. 
Assume now that $ c_i $ is lower semi-continuous. Then the potential function $ P_{\textnormal{FL}} $ is upper semi-continuous over the strategy set $ \mathcal{S} $. Moreover, if $ \mathcal{S} $ is compact, $ P_{\textnormal{FL}} $ attains its maximum on $ \mathcal{S} $. This maximizer corresponds to an NE of the game $ \Gamma_{\textnormal{FL}} $, see \citep[Lem.\@ 2.1]{monderer1996potential}.
\end{proof}

Beyond the existence result, establishing the uniqueness of the NE is generally more challenging, see \Cref{rem_cournot}. A possible condition ensuring uniqueness is the strict concavity of the potential function $ P_{\textnormal{FL}}(s) $. However, this typically requires each cost function $ c_i $ to be strongly convex with a sufficiently large strong convexity modulus. For instance, if $ c_i(x) = \alpha_i x^2 $, then uniqueness holds when $ \lambda < 2\alpha_i/(1+\rho_i^{-1}) $ for all $ i $, which is often impractical in real-world scenarios.
Furthermore, in time-dependent settings (Example~\ref{eg_hetro}), uniqueness is easily lost. See Remark~\ref{rem_non_unique} for further discussion.

\begin{remark}[Comparison to the classic Cournot competition]\label{rem_cournot}
    Note that due to the positivity of $\lambda$, the unit price in \eqref{eq_reward1} increases with the aggregate of clients' efforts. This contrasts with the classic Cournot competition \citep{cournot1927mathematical}, where the price decreases as aggregate effort or production increases. The uniqueness of the NE in this increasing case requires careful analysis (see assumptions of \citep[Thm.\@ 1]{szidarovszky1977new}). 
\end{remark}

In the following section, we investigate the uniqueness of the NE in the stationary and quadratic setting. We show that the NE is unique for all values of $ \lambda $, except at a critical threshold. This result is established via a detailed analysis of the fixed-point system that defines the NE, as presented in Section~\ref{sec_proof}.

\section{Uniqueness Results and the Best-Response Algorithm}\label{sec_homo_game_uniq}

In this section, we focus on the stationary case of $\Gamma_{\textnormal{FL}}$ with quadratic costs $c_i(x)=\alpha_i x^2$. We first reformulate the stationary game, then present uniqueness and threshold results, establish convergence of the best-response algorithm, and finally discuss extensions to non-uniform data sizes and heterogeneous (time-dependent) settings.

\subsection{Stationary Game Formulation}
To simplify the constants in the statements and proofs, we first present the uniform-data-size case, and we discuss the non-uniform counterpart in \Cref{rem_datasize}. Under stationary strategies \eqref{eq:homo_S}, reward rule \eqref{eq_reward1}, and quadratic costs, the game $\Gamma_{\textnormal{FL}}$ is equivalently written as the stationary game $\hat{\Gamma}_{\textnormal{FL}}$ below.

\begin{definition}[Stationary FL game $\hat{\Gamma}_{\textnormal{FL}}$]
\label{def_FL_h_game}
The stationary FL game $\hat{\Gamma}_{\textnormal{FL}}$ consists of the player set (clients) $[m]$, the strategy sets $\hat{\mathcal{S}}_i=[q_i,Q_i]$ for each client, and payoff functions $\hat{P}_i\colon\hat{\mathcal{S}}\to\mathbb{R}$ with $\hat{\mathcal{S}}=\prod_{i\in[m]}\hat{\mathcal{S}}_i$, given by
\begin{equation}\label{game:gamma_h}
    \hat{P}_i(s_i, s_{-i})
    = \frac{\lambda}{m} s_i \sum_{j=1}^m s_j - \alpha_i (s_i)^2, \quad \text{with } \alpha_i>0.
\end{equation}
\end{definition}

{As a special case of $\Gamma_{\textnormal{FL}}$, Theorem~\ref{thm_exist} implies that $\hat{\Gamma}_{\textnormal{FL}}$ is still a potential game and admits NEs.} This stationary formulation is the regime where uniqueness can be characterized sharply, as shown in the next subsection.

\subsection{Uniqueness of the NE in the Stationary Game}\label{subsec_uniq_homo}

Let us introduce the crucial constants {(summarized in Table~\ref{tab_critical_constants})} for the uniqueness result.
\begin{itemize}
\item Concavity threshold:
\begin{equation}\label{b_lambda}
    \bar{\lambda} = m \alpha_{\textnormal{min}}, \, \textnormal{where }\alpha_{\textnormal{min}} \coloneqq \min_{i \in[m]}\{\alpha_i\}.
\end{equation}
\item Jump point $\lambda^* \in (0,\bar{\lambda})$, which is the unique solution to
\begin{equation}\label{eq_lambda_star}
    \sum_{i=1}^m \frac{\lambda^*}{2 m \alpha_i-\lambda^*}=1.
\end{equation}

\item {Two constants $c_1$ and $c_2$ that determine whether a jump occurs at $\lambda^*$:}
\begin{equation}\label{constant_C1_C2}
    c_1 \coloneqq \max_{i \in[m]}\left\{\frac{2 \alpha_i q_i}{\lambda^*}-\frac{q_i}{m}\right\}, \quad
    c_2 \coloneqq \min_{i \in[m]}\left\{\frac{2 \alpha_i Q_i}{\lambda^*}-\frac{Q_i}{m}\right\}.
\end{equation} 
\end{itemize}

\begin{theorem}[Uniqueness]\label{thm_uniq}
The following statements hold:
\begin{enumerate}
    \item If $\lambda >0$ and $\lambda\neq \lambda^{*}$, then the game $\hat{\Gamma}_{\textnormal{FL}}$ has a unique NE.
    \item  If $\lambda = \lambda^{*}$, then the following holds:
    \begin{enumerate}
        \setlength{\itemsep}{1em}
        \item If  $c_1 \geqslant  c_2$, then the game $\hat{\Gamma}_{\textnormal{FL}}$ has a unique NE;
    \item If $c_1 < c_2$, then the game $\hat{\Gamma}_{\textnormal{FL}}$ has infinitely many NEs. A point $s^{*}\in \hat{\mathcal{S}}$ is an NE if and only if there exists $c\in [c_1,c_2]$ such that
        \begin{equation}\label{eq_infinit_NE}
            s^{*}_i = \frac{\lambda^* c}{2 \alpha_i-\lambda^* /m} \,, \quad  \forall i \in [m].
        \end{equation}
    \end{enumerate}
\end{enumerate} 
\end{theorem}
\begin{proof}
The proof is presented in \Cref{sec_proof}.
\end{proof}

\begin{remark}[Role of the concavity threshold $\bar{\lambda}$]
When $\lambda \in (0,\bar{\lambda})$, each payoff $\hat{P}_i(\cdot,s_{-i})$ is concave in $s_i$, so $\hat{\Gamma}_{\textnormal{FL}}$ lies in the classical concave-game framework \citep{rosen1965existence}. Theorem~\ref{thm_uniq}(1) shows that uniqueness can still hold beyond this range. In our analysis, $\bar{\lambda}$ is mainly used to locate the critical value $\lambda^*$ through \eqref{eq_lambda_star}.
\end{remark}

\begin{remark}[A symmetric benchmark]\label{rem_symmetric}
Assume that $\alpha_i=\alpha_j=\alpha$ for all $i,j\in[m]$. Then
\begin{equation}\label{constant_C1_C2_sym}
 \bar{\lambda}=m \alpha, \quad   \lambda^* =\frac{2 m}{m+1} \alpha, \quad
    c_1 = \max _{i \in [m]}\left\{q_i\right\},  \quad
    c_2 = \min _{i \in [m]}\left\{Q_i\right\}.
\end{equation}
In many FL settings, $q_i$ is set close to one local epoch and $Q_i$ is much larger, so $c_1<c_2$ is common. This benchmark also clarifies how the model connects to symmetric mean-field limits \citep[Sec.~2]{cardaliaguet2010notes}.
\end{remark}

\begin{figure}[h]
\centering
\begin{subfigure}[b]{0.31\textwidth}
    \centering
    \includegraphics[width=\textwidth]{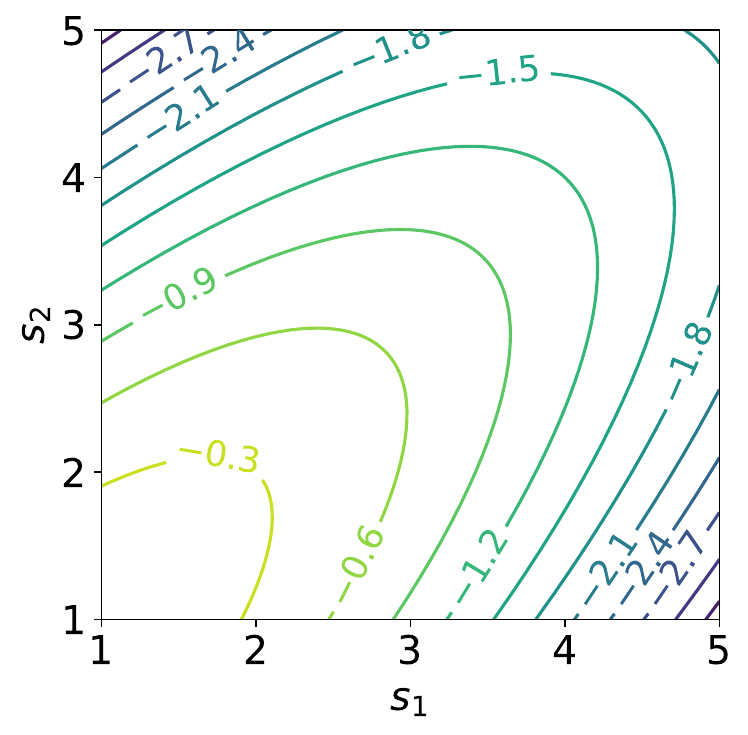}
    \caption{$\lambda < \lambda^*$.}
    \label{fig_contour_a}
\end{subfigure}	
\begin{subfigure}[b]{0.31\textwidth}
    \centering
    \includegraphics[width=\textwidth]{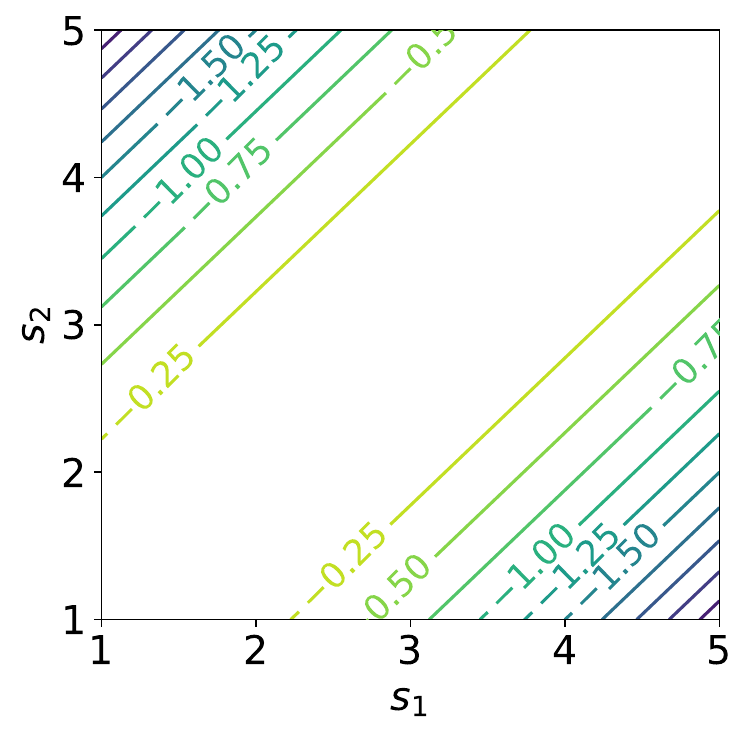}
    \caption{$\lambda = \lambda^*$.}
    \label{fig_contour_b}
\end{subfigure}		
\begin{subfigure}[b]{0.31\textwidth}
    \centering
    \includegraphics[width=\textwidth]{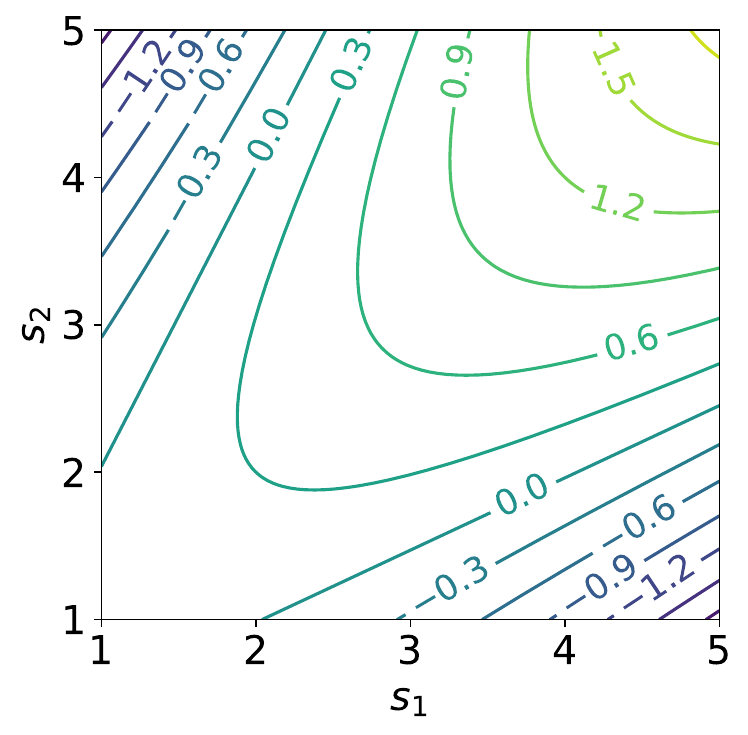}
    \caption{$\lambda > \lambda^*$.}
    \label{fig_contour_c}
\end{subfigure}

\caption{Two-player stationary potential landscape of \Cref{eg_toy2p}. At $\lambda=\lambda^*$, a flat ridge appears along $s_1=s_2$, yielding infinitely many NEs on the feasible diagonal segment. In contrast, for $\lambda<\lambda^*$ and $\lambda>\lambda^*$, the potential has a unique maximizer, corresponding to a unique NE.}
\label{fig_toy2p_landscape}
\end{figure}

\begin{remark}[Geometric interpretation of $\lambda^*$]
\label{rem_geom_lambda_star}
In game $\hat{\Gamma}_{\textnormal{FL}}$, consider the potential
\begin{equation}\label{eq_stationary_potential}
 \hat{P}_{\textnormal{FL}}(s)
 \coloneqq
 \frac{\lambda}{2m}\Big(\sum_{i=1}^m s_i\Big)^2
 + \sum_{i=1}^m\Big(\frac{\lambda}{2m}-\alpha_i\Big)s_i^2,
 \qquad s\in \hat{\mathcal{S}}.
\end{equation}
Its Hessian is
\begin{equation*}\label{eq_stationary_hessian}
 \nabla^2\hat{P}_{\textnormal{FL}}
 =
 \frac{\lambda}{m}\mathbf{1}\mathbf{1}^\top
 + M, \textnormal{ with } M\coloneqq \mathrm{diag}\!\Big(\frac{\lambda}{m}-2\alpha_1,\ldots,\frac{\lambda}{m}-2\alpha_m\Big).
\end{equation*}
For $\lambda<\bar\lambda=m\alpha_{\min}$, the matrix $M$ is nonsingular. By the matrix determinant lemma, 
\[
\det\!\Big(\nabla^2\hat{P}_{\textnormal{FL}}\Big)
=
\det(M)\left(1+\frac{\lambda}{m}\mathbf{1}^\top M^{-1}\mathbf{1}\right).
\]
Therefore, the Hessian is singular if and only if $1+{\lambda}/{m}\sum_{i=1}^m\big({\lambda}/{m}-2\alpha_i\big)^{-1}=0$, that is,
\[
\sum_{i=1}^m \frac{\lambda}{2m\alpha_i-\lambda}=1.
\]
This condition coincides exactly with \eqref{eq_lambda_star}. 
In this case, there exists a nonzero vector \(v\in \mathbb{R}^m\) such that
\[
\nabla^2 \hat{P}_{\textnormal{FL}}\, v = 0.
\]
Moreover, by the homogeneous quadratic structure of \(\hat{P}_{\textnormal{FL}}\), we have
\[
\hat{P}_{\textnormal{FL}}(s+tv)=\hat{P}_{\textnormal{FL}}(s),
\qquad \forall\, s\in\mathbb{R}^m,\ \forall\, t\in\mathbb{R}.
\]
Thus, \(v\) is a flat direction of the potential. If \(s^*\) is a stationary point, then
\[
\nabla \hat{P}_{\textnormal{FL}}(s^*+tv)
=
\nabla \hat{P}_{\textnormal{FL}}(s^*)
+
t\,\nabla^2 \hat{P}_{\textnormal{FL}}\,v
=
0,
\]
so every feasible point on the affine line \(s^*+tv\) is again stationary. Since the stationary potential is concave at \(\lambda=\lambda^*\), these feasible stationary points are global maximizers of \(\hat{P}_{\textnormal{FL}}\). Hence, whenever the feasible set contains a nontrivial segment in this flat direction, the potential admits multiple global maximizers. Since \(\hat{\Gamma}_{\textnormal{FL}}\) is a potential game, these maximizers correspond to NEs, which explains the non-uniqueness of NEs at \(\lambda=\lambda^*\). In the case \(c_1<c_2\), this flat segment connects the low-effort and high-effort branches, which gives the geometric mechanism behind the jump of equilibria across \(\lambda^*\). See \Cref{eg_toy2p} and \Cref{fig_toy2p_landscape} for a visual illustration.

\end{remark}

\begin{example}[Two-player toy landscape]\label{eg_toy2p}
Consider $m=2$ with symmetric parameters $\alpha_1=\alpha_2=1$, $q_1=q_2=1$, and $Q_1=Q_2=5$. Then \eqref{constant_C1_C2_sym} gives $\lambda^*=4/3$, $c_1=1$, and $c_2=5$. Figure~\ref{fig_toy2p_landscape} shows the potential landscape of \eqref{eq_stationary_potential}.
This example illustrates how the critical threshold $\lambda^*$ governs the transition between uniqueness and non-uniqueness of NEs.
\end{example}

\subsection{Nonlinear Threshold Structure of the NE and Jump Phenomenon}
In the following corollary, we present a more precise description of the unique NE of the game $\hat{\Gamma}_{\textnormal{FL}}$ when $\lambda\neq \lambda^{*}$.
Before that, we define two additional critical thresholds, namely the activation point $\lambda_1$ and the saturation point $\lambda_2$ (see \Cref{fig_thresholds}), as follows:
\begin{equation}\label{eq_lambda_1_2}
\lambda_1  \coloneqq 2 \min_{i \in[m]}\left\{\frac{\alpha_i q_i}{\bar{q}+\frac{q_i}{m}}\right\}, \quad
\lambda_2  \coloneqq 2 \, \max_{i \in[m]}\left\{\frac{\alpha_i Q_i}{\bar{Q}+\frac{Q_i}{m}} \right\},
\end{equation}
where $q_i$ and $Q_i$ are the minimum and maximum efforts of client $i \in [m]$, while $\bar{q} = \sum_{i=1}^m q_i / m$ and $\bar{Q} = \sum_{i=1}^m Q_i / m$.
It follows from the definitions that $0 < \lambda_1 \leq \lambda^{*} \leq \lambda_2$.

\begin{corollary}[Precise description of the unique NE]\label{cor_uniq}
The following statements hold:
\begin{enumerate}
    \item If $\lambda \in (0, \lambda_1)$, then the unique NE of the game $\hat{\Gamma}_{\textnormal{FL}}$ is $s^* = (q_i)_{i\in [m]}$.
    \item If $\lambda \in (\lambda_1, \lambda^{*})$, then the unique NE of the game $\hat{\Gamma}_{\textnormal{FL}}$, denoted by $s^{*}$, satisfies
    \begin{equation*}
        \bar{s}^* \coloneqq  \frac{1}{m}\sum_{i\in [m]} s^{*}_i <c_1.
    \end{equation*}
    \item If $\lambda \in (\lambda^{*}, \lambda_2)$, the unique NE of  the game  $\hat{\Gamma}_{\textnormal{FL}}$, denoted by $s^{*}$, satisfies
    \begin{equation*}
        \bar{s}^* \coloneqq \frac{1}{m} \sum_{i\in [m]} s_i^* > c_2.
    \end{equation*}
    \item If  $\lambda \in (\lambda_2, +\infty)$, the unique NE of the game $\hat{\Gamma}_{\textnormal{FL}}$ is $s^* = (Q_i)_{i\in [m]}$.
\end{enumerate}
\end{corollary}
\begin{proof}
The proof is presented in \Cref{sec_proof}.
\end{proof}

\begin{remark}[Regime interpretation]
Corollary~\ref{cor_uniq} yields four regimes, namely inactive incentives $(0,\lambda_1)$, a low-effort branch $(\lambda_1,\lambda^*)$, a high-effort branch $(\lambda^*,\lambda_2)$, and saturation $(\lambda_2,+\infty)$. When $c_1<c_2$, $\lambda^*$ is a genuine jump point. In particular, the unique-NE branches on the two sides of $\lambda^*$ are separated by the gap between $c_1$ and $c_2$, so crossing $\lambda^*$ moves the equilibrium mean effort from at most $c_1$ to at least $c_2$ rather than along a gradual path. This yields a nontrivial jump magnitude of at least $(c_2-c_1)$ in the branch values. The mechanism combines positive reward coupling in \eqref{eq_reward1}, bounded strategy sets \eqref{eq:homo_S}, and the loss of strict curvature at $\lambda^*$ from \Cref{rem_geom_lambda_star}. Consequently, reward factors slightly above $\lambda^*$ form the most sensitive region for incentive tuning.
\end{remark}

\begin{remark}[Robustness beyond the quadratic model]
The closed-form characterization of $\lambda^*$ in \eqref{eq_lambda_star} is specific to the stationary quadratic setting, where best responses are affine and the equilibrium reduces to a scalar fixed-point equation. Beyond this setting, one may still expect qualitatively similar threshold behavior under positively coupled rewards and bounded efforts, but such extensions are not established in the present paper. In particular, linear costs may lead to bang-bang responses and multiplicity, while sufficiently strong convexity may restore uniqueness and smoother dependence on $\lambda$. However, in non-quadratic models, any analogous transition point or jump size would in general have to be characterized implicitly and investigated numerically.
\end{remark}

\subsection{Computation of NEs via the Best-Response Algorithm}\label{subsec_BRA}
The best-response algorithm \citep{durand2016complexity, monderer1996potential, swenson2018best} is a widely employed and easily implemented numerical method for finding an NE in potential games.
Here, we present the best-response algorithm adapted to our framework, as outlined in Algorithm~\ref{alg_BRA}.

\begin{algorithm}
\caption{Best-response algorithm}\label{alg_BRA}
\begin{algorithmic}[1] 
    \State \textbf{Initialization:} $s^{0} \in \hat{\mathcal{S}}$;
    \For{$k = 1, \dots$}
    \For{$i = 1, \dots, m$}
    \If{$s_i^{k-1} \notin \arg\max_{s_i \in \hat{\mathcal{S}}_i} \hat{P}_i(s_1^k, \dots, s_{i-1}^k, s_i, s_{i+1}^{k-1}, \dots, s_m^{k-1})$}
    \State $s_i^k \in \arg\max_{s_i \in \hat{\mathcal{S}}_i} \hat{P}_i(s_1^k, \dots, s_{i-1}^k, s_i, s_{i+1}^{k-1}, \dots, s_m^{k-1})$;
    \Else
    \State $s_i^k = s_i^{k-1}$;
    \EndIf
    \EndFor
    \If{$s^k = s^{k-1}$}
    \State \textbf{return} $s^k$.
    \EndIf
    \EndFor
\end{algorithmic}
\end{algorithm}

A classical convergence result for Algorithm~\ref{alg_BRA}
pertains to finite potential games \citep[Cor.\@ 2.2, Lem.\@ 2.3]{monderer1996potential}. We extend this result to continuous potential games under certain assumptions, as detailed in Assumption~\ref{ass_continuous}. These results offer convergence guarantees for Algorithm~\ref{alg_BRA} when applied to our FL games.

Recall the definitions of $\bar{\lambda}$ and $\lambda^{*}$ from \eqref{b_lambda} and \eqref{eq_lambda_star}. We have the following theorem on the convergence of Algorithm~\ref{alg_BRA} applied to the game $\hat{\Gamma}_{\textnormal{FL}}$.

\begin{theorem}[Convergence of Algorithm~\ref{alg_BRA}]\label{thm_BRA}
Assume that $ \lambda \in (0,\bar{\lambda})$ and $ \lambda \neq \lambda^{*}$. For $k \geqslant 1$, let $s^k$ be the result of Algorithm~\ref{alg_BRA} at the $k$-th iteration. Then, the sequence $\{s^k\}_{k\geqslant 1}$ converges to the unique NE of $\hat{\Gamma}_{\textnormal{FL}}$.
\end{theorem}

\begin{proof}
The game $\hat{\Gamma}_{\textnormal{FL}}$ satisfies Assumption~\ref{ass_continuous}. By applying the general convergence result of the best-response algorithm from Theorem~\ref{thm_BRA_general}(2), we obtain that any limit point of the sequence $\{s^k\}_{k\geqslant 1}$ is an NE of $\hat{\Gamma}_{\textnormal{FL}}$. By Theorem~\ref{thm_uniq}, if $\lambda \neq \lambda^{*}$, this NE is unique, which implies that the limit point of $\{s^k\}_{k\geqslant 1}$ is unique. Combining this with the pre-compactness of $\{s^k\}_{k\geqslant 1}$, the conclusion follows.
\end{proof}

Building on Theorem~\ref{thm_BRA_general}, the following remark establishes the convergence of Algorithm~\ref{alg_BRA} for the specific case $\lambda = \lambda^*$, serving as a complement to Theorem~\ref{thm_BRA}.
\begin{remark}\label{rem_algo_uniq}
Recall the definitions of $c_1$ and $c_2$ from \eqref{constant_C1_C2}. If $c_1 \geqslant c_2$, the convergence result in Theorem~\ref{thm_BRA} holds for $\lambda=\lambda^{*}$ (see \Cref{thm_uniq}(2a)). Additionally, if $c_1 < c_2$, at the point $\lambda = \lambda^{*}$, any cluster point of the sequence $\{s^k\}_{k\geqslant 1}$ is an NE of $\hat{\Gamma}_{\textnormal{FL}}$. In all these cases, for any $K \geqslant 1$, there exists $k \in \{1, \ldots, K\}$ such that $s^k$ is an $\mathcal{O}(1/K)$-NE of $\hat{\Gamma}_{\textnormal{FL}}$. Here, we refer to Definition~\ref{def:eNE} for the definition of an $\epsilon$-NE.
\end{remark}

\subsection{Discussion of Game Models}
In this subsection, we discuss both theoretical extensions and the practical interpretation of the FL game. On the theoretical side, we consider non-uniform data sizes and time-dependent (round-heterogeneous) strategies. On the practical side, we clarify how $\alpha_i$, $q_i$, and $Q_i$ can be calibrated and how the model should be interpreted as a reduced-form incentive mechanism, with data-dependent effects incorporated implicitly through parameters.

\begin{remark}[{Non-uniform data sizes}]\label{rem_datasize}
In Definition~\ref{def_FL_h_game}, we assume that all clients have equal data sizes to simplify the constants and the analysis in Theorem~\ref{thm_uniq}.
In the non-uniform case, the existence of an NE is guaranteed by Theorem~\ref{thm_exist}. Regarding uniqueness, by following the proof in \Cref{sec_proof}, we can establish the same results as in Theorem~\ref{thm_uniq} and Corollary~\ref{cor_uniq}, with the constants introduced in \eqref{b_lambda}-\eqref{constant_C1_C2} and \eqref{eq_lambda_1_2} replaced by
\begin{align*}
    & \bar{\lambda}  = \min_{i\in [m]}\{ \alpha_i / \rho_i \} ; \quad \lambda^* \in (0, \bar{\lambda}), \textnormal{ s.t. } \sum_{i=1}^m \frac{\lambda^*}{2 \alpha_i / \rho_i-\lambda^*}=1;\\
    & c_1 = \max_{i \in[m]}\left\{\frac{2 \alpha_i q_i}{\lambda^*}- \rho_i q_i \right\}, \quad
    c_2 = \min_{i \in[m]}\left\{\frac{2 \alpha_i Q_i}{\lambda^*}-\rho_i Q_i \right\}; \\
    & \lambda_1  = 2 \min_{i \in[m]}\left\{\frac{\alpha_i q_i}{\bar{q}+\rho_i q_i }\right\}, \quad
\lambda_2  = 2 \, \max_{i \in[m]}\left\{\frac{\alpha_i Q_i}{\bar{Q}+\rho_i Q_i} \right\},
\end{align*}
where $\bar{q} = \sum_{i\in [m]} \rho_i q_i $ and $\bar{Q} = \sum_{i\in [m]} \rho_i Q_i $. Note that if $\rho_i = 1/m$ for all $i\in [m]$ (\textit{i.e.}, the uniform case), then we recover the constants defined in Section~\ref{subsec_uniq_homo}. In addition, we obtain the same results as in Theorem~\ref{thm_BRA} and Remark~\ref{rem_algo_uniq} for the convergence of Algorithm~\ref{alg_BRA} in this non-uniform case.
\end{remark}

\begin{remark}[Time-dependent games]\label{rem_non_unique}
A key assumption for the uniqueness results presented in this section is the stationarity of training efforts with respect to the communication rounds, \textit{i.e.}, $s_i^t$ is independent of $t$. However, in the time-dependent case (the original game $\Gamma_{\text{FL}}$ with strategy sets \eqref{eq_hetro_S}), uniqueness can easily be lost. For instance, suppose that there exists a time-dependent NE $s^{*}$ with rounds $t_1, t_2 \in [T]$ such that $(s^{*})^{t_1} \neq (s^{*})^{t_2}$. Then, from the potential function \eqref{potential_func}, we can verify that $\tilde{s}^*$ (which differs from $s^*$) is also an NE of the game $\Gamma_{\text{FL}}$, where $\tilde{s}^*$ is obtained by swapping the training efforts of $s^*$ at rounds $t_1$ and $t_2$.

Although the uniqueness result does not hold, Algorithm~\ref{alg_BRA} still converges in the time-dependent case, as in Remark~\ref{rem_algo_uniq}. Let $s^k$ denote the outcome of iteration $k$ in Algorithm~\ref{alg_BRA} adapted to $\Gamma_{\text{FL}}$. Then, any cluster point of $s^k$ is an NE of $\Gamma_{\text{FL}}$.
\end{remark}

\begin{remark}[Practical calibration and interpretation]\label{rem_data_agnostic}
In practice, the server can calibrate $\alpha_i$, $q_i$, and $Q_i$ via device profiling, historical participation logs, or lightweight surveys. At the modeling level, however, we emphasize that the proposed game is a reduced-form incentive model. Equilibrium efforts are determined by cost--reward considerations and do not explicitly encode the full local data distribution or the detailed state of the global model. This decoupling is deliberate, because it preserves a transparent potential-game structure while avoiding direct access to private data statistics.

Data-dependent effects can still be incorporated implicitly through model parameters. In the baseline design, $\rho_i$ is tied to relative data size, but it can also be interpreted more broadly as a contribution weight that reflects data quality or usefulness. Likewise, $\alpha_i$ can represent an effective participation cost, including computation, energy, and opportunity costs associated with local data value. One can also introduce model-state dependence by multiplying \eqref{eq_reward1} by a round-dependent factor, for example $\gamma_t=1/\sqrt{t}$, so that incentives are stronger in early rounds and weaker in later rounds.

Therefore, the resulting NE should be interpreted as a tractable proxy for local computation rather than as a closed-form predictor of final model accuracy. Our theory characterizes how incentives shift equilibrium effort. The induced reward--performance relation is then assessed empirically in Section~\ref{sec_numrical}. Developing fully data-aware reward rules remains a natural extension (see Section~\ref{sec_conclusions}).
\end{remark}

\section{Numerical Experiments}\label{sec_numrical}
In this section, we present numerical experiments for the FL game and the induced FL training process. We first describe the experimental setup. We then investigate the NEs (\emph{i.e.}, the clients' rational training efforts) of the stationary game $\hat{\Gamma}_{\textnormal{FL}}$ under different reward factors $\lambda$. Finally, we incorporate the resulting equilibrium efforts into FL training on different datasets and models, thereby validating the practical relevance of the proposed game-theoretic framework. The code for reproducing all experiments is available at {\small\url{https://github.com/DCN-FAU-AvH/FL-Potential-Game}.}

\subsection{Experimental Setup}\label{subsec_simu_param}
\paragraph{FL game scenarios}
We consider a stationary FL game with $m \in \{1,000,3,000\}$ clients over $T=200$ rounds. The server selects a reward factor $\lambda \in [0,5]$, and each client $i$ has a local cost coefficient $\alpha_i \in [1,2]$. We set the training effort $s_i$ to be the number of local epochs, with a fixed minimum effort $q_i = 1$ and a maximum feasible effort $Q_i \in [20,30]$. This choice is representative of FL practice, where at least one local epoch is required once a client participates. The corresponding thresholds $\lambda_1$, $\lambda^*$, and $\lambda_2$ are summarized in \Cref{tab_cases}.

\paragraph{Datasets and algorithms}
We consider representative FL tasks, including image classification on \emph{MNIST} \citep{lecun1998gradientbased}, \emph{CIFAR-10}, and \emph{CIFAR-100} \citep{krizhevsky2009learning}, sentiment analysis on \emph{IMDB} \citep{maas2011learning}, and character recognition on \emph{FEMNIST} \citep{caldas2018leaf}. Training is conducted using two widely adopted FL algorithms, namely FedAvg \citep{mcmahan2017communicationefficient} and FedProx \citep{li2020federateda}.

\paragraph{Data partitioning and training}
For \emph{CIFAR-10}, \emph{CIFAR-100}, and \emph{IMDB}, the data are uniformly distributed across 1,000 clients, and 10\% of the clients are sampled in each round. To induce strong label skew, each client is restricted to data from at most two labels, which yields a pathological non-IID distribution. We use a batch size of 20 and a learning rate of 0.05.
For \emph{FEMNIST}, we consider 3,000 clients and sample 1\% of the clients in each round. This dataset is inherently user-partitioned, naturally non-IID, and highly imbalanced. We use a batch size of 50 and a learning rate of 0.001 for this dataset.

\paragraph{Neural network architectures}
For image classification, we use a convolutional neural network with two convolutional layers of 32 and 64 channels, followed by a fully connected layer of size 512, following \citet{mcmahan2017communicationefficient}. For sentiment analysis, we use an encoder-only Transformer with a 128-dimensional embedding layer and two encoder layers, each equipped with 2-head self-attention and a feedforward sublayer of size 128.

\paragraph{Hardware} All experiments are conducted on a server equipped with an AMD EPYC 7662 processor, 64 GB RAM, and an NVIDIA A100 GPU.

\subsection{Evolution of NEs and Critical Reward Factors}\label{subsec_simu_NE}

\begin{figure}[t]
\centering
\begin{subfigure}[b]{0.45\textwidth}
    \centering
    \includegraphics[width=\textwidth]{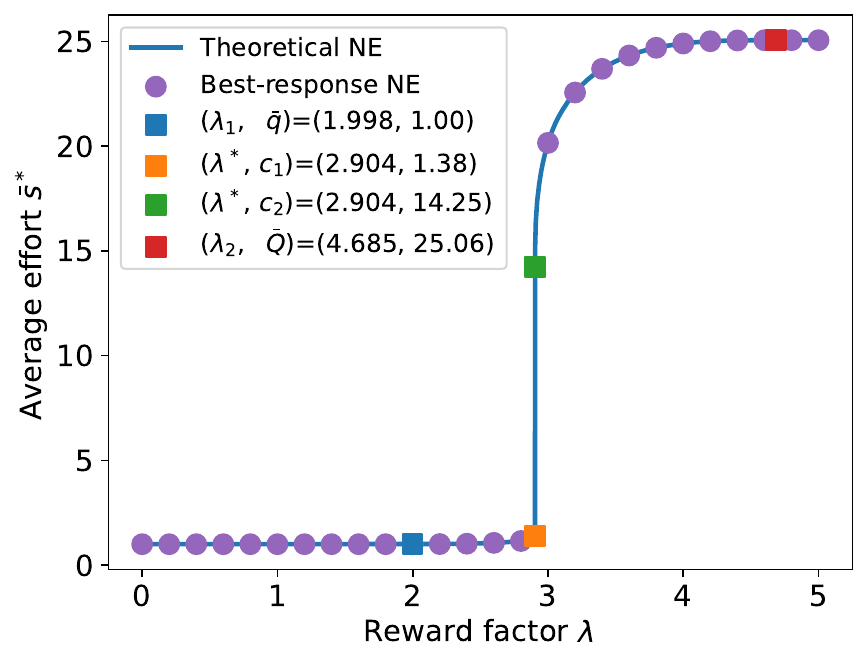}
        \caption{$m=1,000$ clients.}
    \label{fig_NEs_a}
\end{subfigure}%
\begin{subfigure}[b]{0.45\textwidth}
    \centering
    \includegraphics[width=\textwidth]{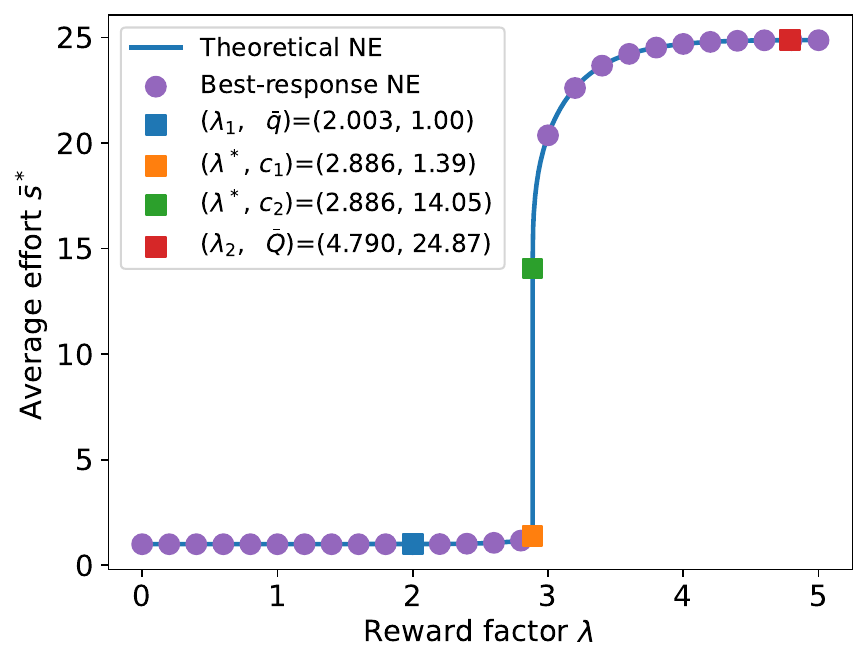}
        \caption{$m=3,000$ clients.}
    \label{fig_NEs_c}
\end{subfigure}%

\begin{subfigure}[b]{0.31\textwidth}
    \centering
    \includegraphics[width=\textwidth]{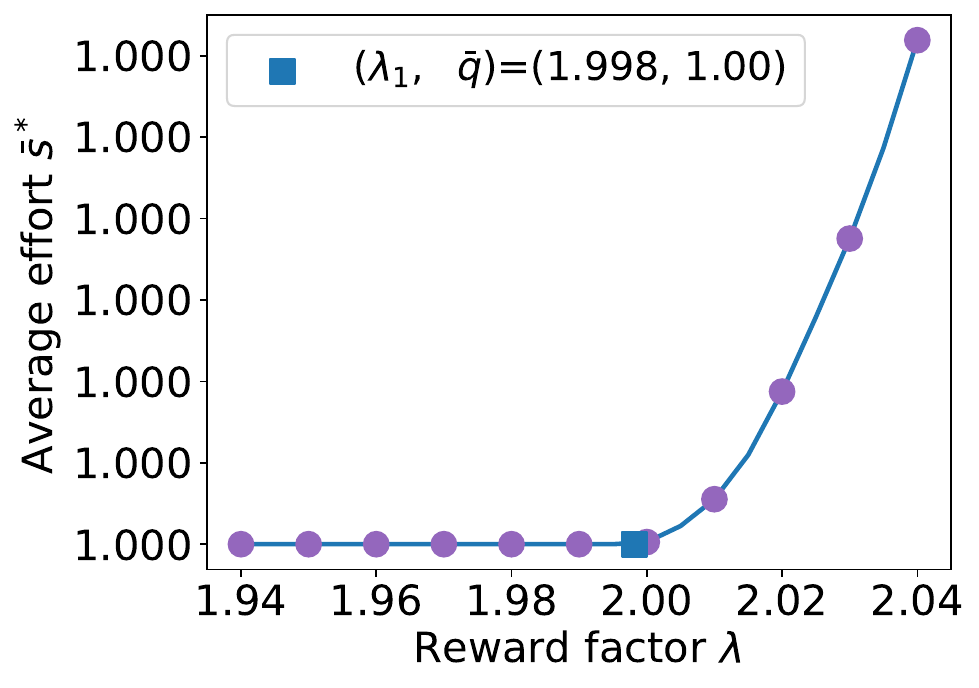}
    \caption{$m=1,000$, zoom-in at $\lambda_1$.}
    \label{fig_NE_zoom_in_a}
\end{subfigure}	
\begin{subfigure}[b]{0.29\textwidth}
    \centering
    \includegraphics[width=\textwidth]{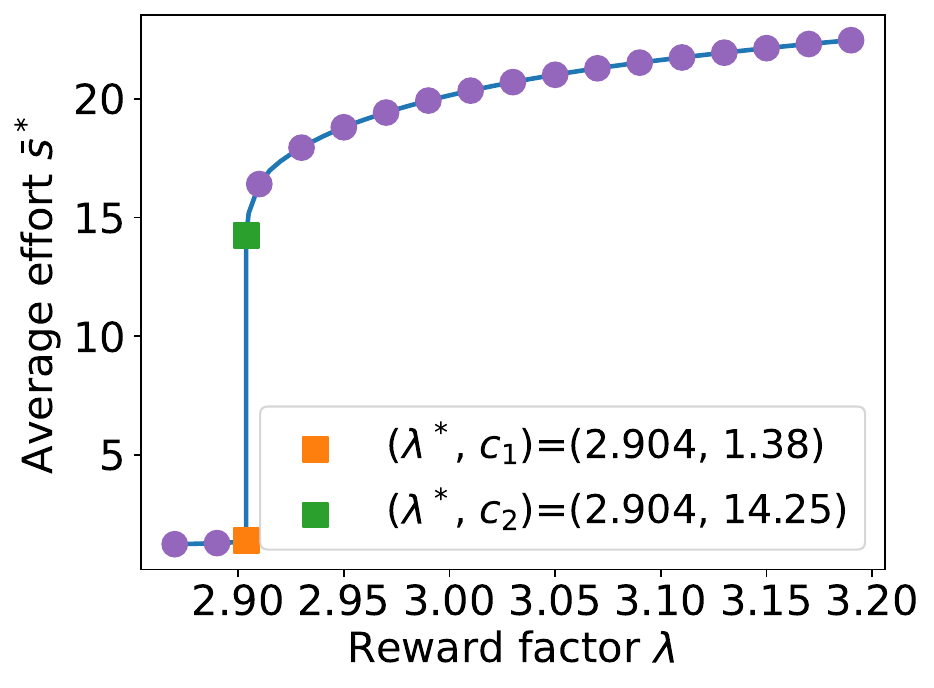}
    \caption{$m=1,000$, zoom-in at $\lambda^*$.}
    \label{fig_NE_zoom_in_b}
\end{subfigure}		
\begin{subfigure}[b]{0.31\textwidth}
    \centering
    \includegraphics[width=\textwidth]{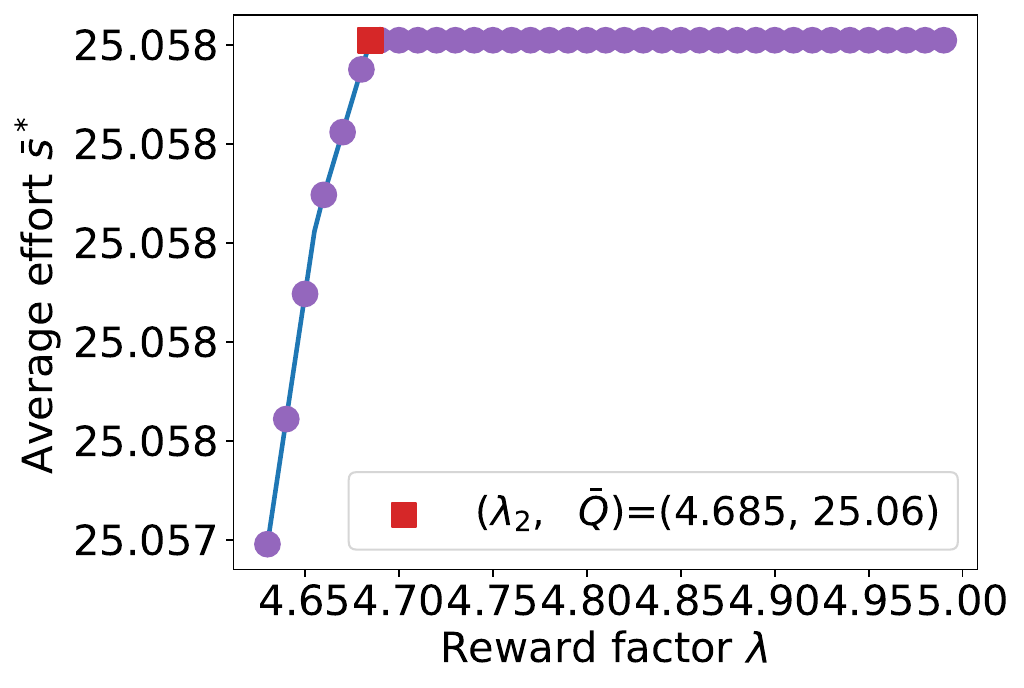}
    \caption{$m=1,000$, zoom-in at $\lambda_2$.}
    \label{fig_NE_zoom_in_c}
\end{subfigure}
\caption{Average training effort $\bar{s}^*$ as a function of the server's reward factor $\lambda$. The activation point $\lambda_1$, jump point $\lambda^*$, and saturation point $\lambda_2$ are indicated in the legends.}
\label{fig_NEs}
\end{figure}

In this subsection, we validate the theoretical results in \Cref{subsec_uniq_homo} by examining how the average effort $\bar{s}^*$ varies with the server's reward factor $\lambda$. The NEs are computed using Algorithm~\ref{alg_BRA} and are compared with the theoretical values derived from the fixed-point system \eqref{eq_fixed_point}. The results shown in \Cref{fig_NEs} indicate that the equilibria obtained from the theoretical characterization and the best-response method are indistinguishable up to numerical precision, thereby validating both the fixed-point characterization and the best-response solver.

Furthermore, as shown in \Cref{fig_NEs}, the average effort $\bar{s}^*$ increases monotonically with $\lambda$ for both $m=1{,}000$ and $m=3{,}000$ clients. In the early regime with $\lambda < \lambda_1$, $\bar{s}^*$ remains at its minimum value $\bar{q}$, as shown in \Cref{fig_NE_zoom_in_a}. In the late regime with $\lambda > \lambda_2$, it saturates at the maximum value $\bar{Q}$, as shown in \Cref{fig_NE_zoom_in_c}. These observations are consistent with \Cref{cor_uniq} and suggest that the practically relevant range of $\lambda$ lies in $(\lambda_1,\lambda_2)$, where the model avoids both under-training and saturation.

For the intermediate regime with $\lambda \in (\lambda_1,\lambda_2)$, \Cref{fig_NE_zoom_in_b} shows that $\bar{s}^*$ exhibits a pronounced jump from $c_1$ to $c_2$ at $\lambda^*$. According to \Cref{thm_uniq}(2b), every point between $c_1$ and $c_2$ corresponds to an NE\@. This observation indicates that reward factors slightly above $\lambda^*$ are especially effective at stimulating collective effort. We examine this implication further in the FL training experiments of the next subsection. Moreover, the jump at $\lambda^*$ is observed consistently in both the 1,000-client setting in \Cref{fig_NEs_a} and the 3,000-client setting in \Cref{fig_NEs_c}, which suggests that the theoretical prediction remains robust as the number of clients increases.

Overall, these results highlight the strong sensitivity of the NEs, namely the clients' rational training efforts, to the server's reward factor. To connect the above equilibrium analysis with FL training, we next select four representative equilibrium cases around the thresholds $\lambda_1$, $\lambda^*$, and $\lambda_2$, and apply the corresponding effort profiles to practical FL tasks to evaluate their effectiveness.

\subsection{Training Performance of FL under Various NEs}\label{subsec_simu_FL_train}
In this subsection, we evaluate how different NEs affect FL training performance across benchmark datasets, models, and algorithms.
We emphasize that the proposed equilibrium mechanism is not a new FL algorithm that prescribes clients' training efforts. Rather, it is a modeling framework for the incentive layer. Specifically, when rewards are absent or weak, clients may have limited willingness to participate actively, and the game is used to characterize their rational effort responses under server-side incentives. Accordingly, the goal of this subsection is to assess the practical relevance of this modeling framework by examining how the resulting equilibrium effort profiles affect FL training performance.

We begin with four representative NE cases located near the three critical thresholds $\lambda_1$, $\lambda^*$, and $\lambda_2$. Specifically, Case~1 is chosen before the activation point $\lambda_1$. Case~2 is chosen immediately before the jump point $\lambda^*$. Case~3 is chosen immediately after the jump point. Case~4 is chosen beyond the saturation point $\lambda_2$. We then apply the equilibrium efforts $(s_i^*)_{i\in[m]}$ under these four cases to various FL settings. In all experiments, each client performs $\lceil s_i^* \rceil$ epochs of local training, where $\lceil \cdot \rceil$ denotes the ceiling function.

\begin{figure}[t]
\centering
\includegraphics[width=\textwidth]{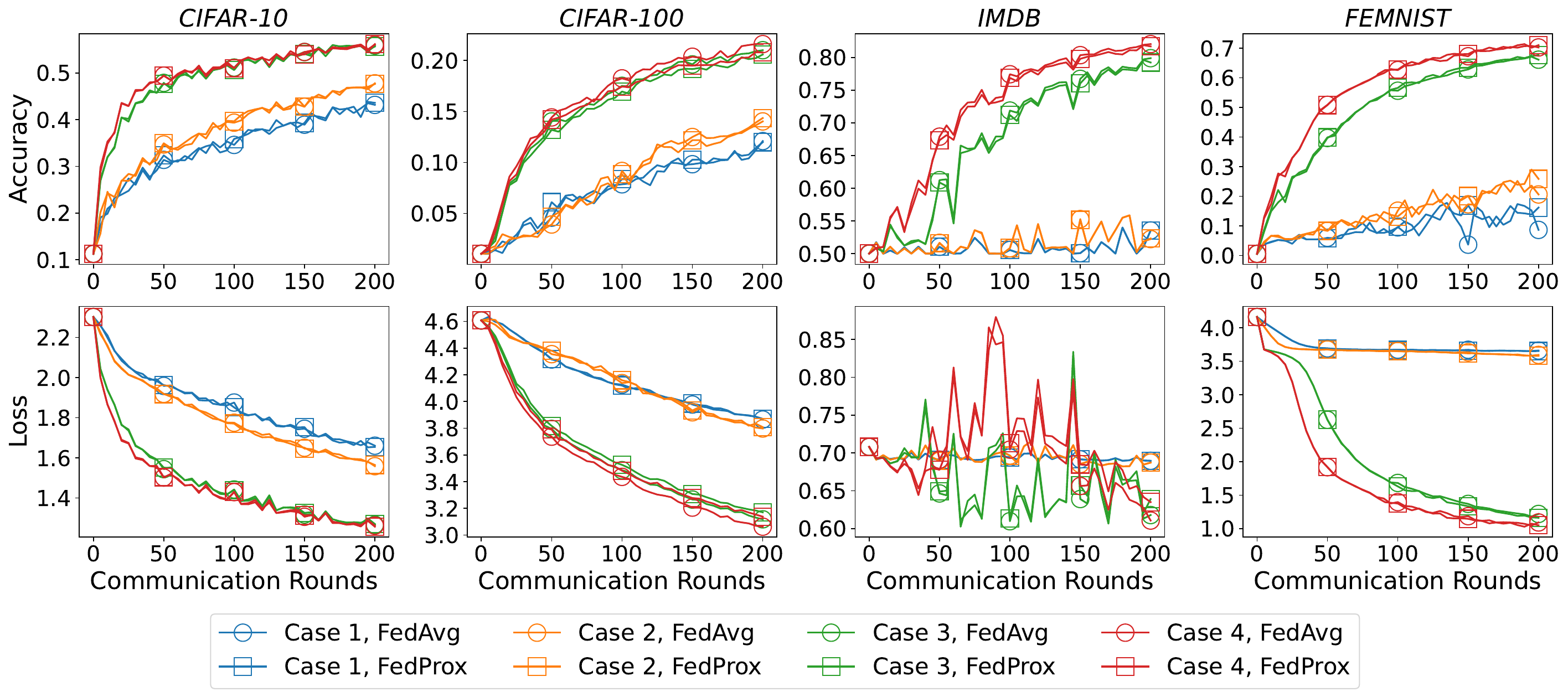}
\caption{Results of FL training under four NE cases across different datasets and algorithms. The largest gain is observed when moving from Case~2 to Case~3 in all experiments.}
\label{fig_FL_train}
\end{figure}

The specific values of $\lambda$, the average efforts $\bar{s}^*$, and the test results for the four cases are summarized in \Cref{tab_cases} and \Cref{fig_FL_train}. The four cases exhibit a broadly consistent pattern across the experiments. The most substantial improvement occurs when moving from Case~2 to Case~3, which underscores the critical role of the jump point $\lambda^*$.

The main observations are summarized below.
\begin{enumerate}
    \item \textbf{Case 1, selected before the activation point $\lambda_1$.} This case yields the weakest performance. Both the accuracy and loss curves indicate insufficient training, which is consistent with the low effort level of the clients.
    \item \textbf{Case 2, selected before the jump point $\lambda^*$.} Performance remains close to that of Case~1. The gain is limited overall, which suggests that the modest increase in equilibrium effort is still insufficient to produce a clear improvement in training quality.
    \item \textbf{Case 3, selected after the jump point $\lambda^*$.} This case shows a marked improvement in training performance, with a clear increase in accuracy and a simultaneous reduction in loss. This pattern appears consistently across the tasks and supports the practical relevance of the theoretical jump at $\lambda^*$.
    \item \textbf{Case 4, selected after the saturation point $\lambda_2$.} Further increases in training effort yield only marginal additional gains in accuracy and only slight reductions in loss. This behavior indicates diminishing returns once the reward factor has passed the jump region.
\end{enumerate}

\begin{table}[t]
\centering
\caption{Summary of experimental cases. Cases 1--4 correspond, respectively, to the pre-activation, pre-jump, post-jump, and post-saturation regimes, that is, before $\lambda_1$, immediately before $\lambda^*$, immediately after $\lambda^*$, and after $\lambda_2$. Test accuracies are averaged across different FL algorithms.}
\label{tab_cases}
\small
\setlength{\tabcolsep}{4pt}
\resizebox{\textwidth}{!}{%
\begin{tabular}{clcccccccc}
\toprule
 &  & \multicolumn{5}{c}{Balanced, non-IID, $m=1{,}000$} & \multicolumn{3}{c}{Imbalanced, non-IID, $m=3{,}000$} \\
\cmidrule(lr){3-7}\cmidrule(l){8-10}
 &  & \multicolumn{5}{c}{$\lambda_1 = 1.998,\ \lambda^* = 2.904,\ \lambda_2 = 4.685$} & \multicolumn{3}{c}{$\lambda_1 = 2.003,\ \lambda^* = 2.886,\ \lambda_2 = 4.790$} \\
\cmidrule(lr){3-7}\cmidrule(l){8-10}
 &  & \multicolumn{2}{c}{NE} & \multicolumn{3}{c}{Test accuracy} & \multicolumn{2}{c}{NE} & Test accuracy \\
\cmidrule(lr){3-4}\cmidrule(lr){5-7}\cmidrule(l){8-9}\cmidrule(l){10-10}
Case & Regime & $\lambda$ & $\bar{s}^*$ & CIFAR-10 & CIFAR-100 & IMDB & $\lambda$ & $\bar{s}^*$ & FEMNIST \\
\midrule
Case 1 & pre-activation & 1.99 & 1.00 & 0.434 & 0.120 & 0.535 & 2.00 & 1.00 & 0.124 \\
Case 2 & pre-jump & 2.90 & 1.32 & 0.477 & 0.142 & 0.523 & 2.88 & 1.32 & 0.232 \\
Case 3 & post-jump & 2.91 & 16.41 & 0.558 & 0.208 & 0.796 & 2.89 & 15.94 & 0.665 \\
Case 4 & post-saturation & 4.69 & 25.05 & 0.561 & 0.212 & 0.819 & 4.80 & 24.87 & 0.706 \\
\bottomrule
\end{tabular}
}
\end{table}

Overall, these results suggest that increasing the reward factor $\lambda$ can improve training performance, but the most substantial gains occur around the jump point $\lambda^*$. Beyond this region, additional incentives provided by the server lead only to limited further improvement. If the goal is to obtain a large performance gain without moving deep into the saturation regime, then reward factors slightly above $\lambda^*$ are natural candidates.

\section{Technical Proofs and Theorems}\label{sec_proof}
This section presents the detailed proofs of all theoretical results stated in the paper.
We first rewrite the NE conditions as an explicit best-response system and reduce the
$m$-dimensional equilibrium problem to a scalar fixed-point equation for the mean effort.
We then study how this scalar equation depends on the reward factor $\lambda$.
We finally establish convergence guarantees for the best-response algorithm in potential games.

\subsection{Proofs of Theorem~\ref{thm_uniq} and Corollary~\ref{cor_uniq}}\label{subsec_proof_part1}
Recall the parameters $\lambda$, $m$, $\alpha_i$, $q_i$, and $Q_i$ of the game $\hat{\Gamma}_{\textnormal{FL}}$ from Definition~\ref{def_FL_h_game} and Example~\ref{eg_homo}. Recall also the payoff function $\hat{P}_i$ in \eqref{game:gamma_h}. The next lemma derives the fixed-point conditions satisfied by the NEs of $\hat{\Gamma}_{\textnormal{FL}}$ from the quadratic form of $\hat{P}_i$.
\begin{lemma}\label{NE:fixed_point_1}
A point $s^{*}\in \prod_{i\in[m]} [q_i,Q_i]$ is an NE of the game $\hat{\Gamma}_{\textnormal{FL}}$ if and only if the following holds:
\begin{subequations}
\begin{empheq}[left={s_i^* = \empheqlbrace}]{align}
    q_i \qquad , \quad &\textnormal{if} \ q_i \geqslant \frac{\lambda\left(\bar{s}^*-s^*_i/m\right)}{2\left(\alpha_i-\lambda/m\right)} \,\textnormal{ and }\, \alpha_i > \frac{\lambda}{m} ,\label{eq_s_i-1} 
    \\[0.6em]
    \frac{\lambda \bar{s}^*}{2 \alpha_i-\lambda/m}, \quad &\textnormal{if} \ q_i<\frac{\lambda\left(\bar{s}^*-s^*_i/m\right)}{2\left(\alpha_i-\lambda/m\right)}<Q_i, \label{eq_s_i-2}
    \\[0.6em]
    Q_i \quad \,\,\,\, , \quad  &\textnormal{if} \ Q_i \leqslant \frac{\lambda\left(\bar{s}^*-s^*_i/m\right)}{2\left(\alpha_i-\lambda/m\right)}\,\textnormal{ or }\, \alpha_i \leqslant \frac{\lambda}{m}, \label{eq_s_i-3}
\end{empheq}
\end{subequations}
where $ \bar{s}^* = \sum_{i\in[m]} s^{*}_i/m$.
\end{lemma}

The next lemma provides an equivalent formulation of \eqref{eq_s_i-1}-\eqref{eq_s_i-3}, which will be used in the proof of Theorem~\ref{thm_uniq}. To this end, for each $i \in [m]$, define the auxiliary mapping $\beta_i \colon \mathbb{R}_{+}\to \mathbb{R}_{+}$ by
\begin{subequations}
\begin{empheq}[left={\beta_i (x) \coloneqq \empheqlbrace}]{align}
    q_i \qquad, \quad &\textnormal{if} \ q_i \geqslant \frac{\lambda \,x }{2 \alpha_i-\lambda/m} \,\textnormal{ and }\, \alpha_i > \frac{\lambda}{m} ,\label{eq_beta_i-1}
    \\[0.6em]
    \frac{\lambda x}{2 \alpha_i-\lambda/m}, \quad &\textnormal{if} \ q_i < \frac{\lambda  \, x}{2 \alpha_i-\lambda/m} < Q_i  , \label{eq_beta_i-2}
    \\[0.6em]
    Q_i \quad \,\,\,\, , \quad  &\textnormal{if} \  Q_i \leqslant \frac{\lambda \, x}{2 \alpha_i-\lambda/m}\,\textnormal{ or }\, \alpha_i \leqslant \frac{\lambda}{m}.\label{eq_beta_i-3}
\end{empheq}
\end{subequations}

\begin{lemma}\label{NE:fixed_point}
Let $s^{*}$ be an NE of the game $\hat{\Gamma}_{\textnormal{FL}}$ and let $  \bar{s}^* = \sum_{i\in [m]} s^*_i / m$. Then, it holds:
\begin{equation}\label{eq_best_response}
    s_i^*  = \beta_i(\bar{s}^*), \quad \textnormal{for } i\in [m].
\end{equation}
As a consequence, $\bar{s}^{*}$ satisfies the following fixed-point equation:
\begin{equation}\label{eq_fixed_point}
    \bar{s}^* = \frac{1}{m}\sum_{i\in [m]} \beta_i(\bar{s}^*).
\end{equation}
\end{lemma}
\begin{proof}
By Lemma~\ref{NE:fixed_point_1}, $s^*$ and $\bar{s}^*$ satisfy \eqref{eq_s_i-1}-\eqref{eq_s_i-3}. For $i\in[m]$ such that $s^{*}_i = q_i$, it follows from \eqref{eq_s_i-1} that
\begin{equation*}
    q_i \geqslant \frac{\lambda\left(\bar{s}^*-q_i/m\right)}{2\left(\alpha_i-\lambda/m\right)} \,\text{ and }\, \alpha_i > \frac{\lambda}{m},
\end{equation*}
which is equivalent to \eqref{eq_beta_i-1} by a direct calculation. This implies that $s_{i}^* = \beta_i(\bar{s}^*)$ for these values of $i$.
By a similar argument, we obtain that $s_{i}^* = \beta_i(\bar{s}^*)$ for $i\in[m]$ such that $s^{*}_i = Q_i$. For $i\in[m]$ such that $s^{*}_i = \lambda \bar{s}^*/(2 \alpha_i- \lambda/m)$, the equality
\begin{equation*}
    \frac{\lambda\left(\bar{s}^*-s_i^*/m\right)}{2\left(\alpha_i-\lambda/m\right)} = \frac{\lambda \bar{s}^*}{2 \alpha_i-\lambda/m},
\end{equation*}
implies the equivalence between \eqref{eq_s_i-2} and \eqref{eq_beta_i-2}, which gives $s_{i}^* = \beta_i(\bar{s}^*)$ in this case.
\end{proof}

From \eqref{eq_beta_i-1}-\eqref{eq_beta_i-3}, we define the following four set-valued mappings on $\mathbb{R}_+$:
\begin{align*}
I_1(x) &\coloneqq \left\{ i \in [m]\, \Big|\, q_i \geqslant \frac{\lambda \,x }{2 \alpha_i-\lambda/m} \,\textnormal{ and }\, \alpha_i > \frac{\lambda}{m} \right\};\\[0.6em]
I_2(x) &\coloneqq \left\{ i \in [m]\, \Big|\, q_i < \frac{\lambda  \, x}{2 \alpha_i-\lambda/m} < Q_i \right\};\\[0.6em]
I_3(x) &\coloneqq \left\{ i \in [m]\, \Big|\, Q_i \leqslant \frac{\lambda \, x}{2 \alpha_i-\lambda/m}\,\textnormal{ and }\, \alpha_i > \frac{\lambda}{m} \right\};\\[0.6em]
I_4(x) &\coloneqq \left\{ i \in [m]\, \Big|\,  \alpha_i \leqslant \frac{\lambda}{m}  \right\}.
\end{align*}
By construction, the conditions in \eqref{eq_beta_i-1}-\eqref{eq_beta_i-3} are equivalent to $i \in I_1(x)$, $i\in I_2(x)$, and $i\in  I_3(x) \cup I_4(x)$, respectively.

\begin{lemma}
Fix any $x\in \mathbb{R}_+$. The following statements hold:
\begin{enumerate}
    \item The sets $I_1(x)$, $I_2(x)$, $I_3(x)$, and $I_4(x)$ are mutually disjoint, and
    \begin{equation}\label{eq_union}
        I_1(x) \cup I_2(x) \cup I_3(x)\cup I_4(x) = [m].
    \end{equation}
    \item For any $y \in \mathbb{R}_+$ such that $x \leq y$, we have
    \begin{equation}\label{eq_inclusion}
        I_1(y) \subseteq I_1(x), \quad \textnormal{and} \quad I_3(x) \subseteq I_3(y).
    \end{equation}
\end{enumerate}
\end{lemma}
\begin{proof}
The proof is straightforward by the definitions of $I_1$, $I_2$, $I_3$, and $I_4$.
\end{proof}

\begin{proof}[Proof of Theorem~\ref{thm_uniq}.]
By Lemma~\ref{NE:fixed_point}, every equilibrium is determined by the scalar mean effort
$\bar{s}$ together with the partition $I_1$, $I_2$, $I_3$, and $I_4$.
Assume that two distinct equilibria exist and let their means satisfy $\bar{s}^1<\bar{s}^2$.
If $\lambda\neq\lambda^*$, the monotonicity of the index sets and the two fixed-point identities
yield incompatible bounds on $\sum_{i} \lambda/(2m\alpha_i-\lambda)$, which gives a contradiction.
If $\lambda=\lambda^*$, the same reduction shows that non-uniqueness can only occur on the interior
affine branch $s_i=\lambda^* c/(2\alpha_i-\lambda^*/m)$, and the comparison between $c_1$ and $c_2$
distinguishes the unique case from the continuum case.

\noindent\textbf{Step 1} (The case $\lambda\in (0,\bar{\lambda})$ and $\lambda\neq \lambda^*$).   
Assume that there exist two distinct NEs, denoted by $s^1$ and $s^2$. Let $\bar{s}^1 = \sum_{i\in [m]} s^1_i/m$ and $\bar{s}^2 = \sum_{i\in [m]} s^2_i/m$. By \eqref{eq_best_response}, we have $s^1_i = \beta_i(\bar{s}^1)$ and $s^2_i = \beta_i(\bar{s}^2)$ for all $i\in [m]$. Since $s^1 \neq s^2$, we have $\bar{s}^1 \neq \bar{s}^2$. Without loss of generality, assume that $\bar{s}^1 < \bar{s}^2$. By the definition of $\bar{\lambda}$ in \eqref{b_lambda} and the inequality $\lambda < \bar{\lambda}$, we have $I_4(\bar{s}^1) = I_4(\bar{s}^2) = \emptyset$.
Then \eqref{eq_fixed_point}, \eqref{eq_union}, and \eqref{eq_beta_i-1}-\eqref{eq_beta_i-3} imply
\begin{align}
    m \bar{s}^1 & = \sum_{i\in I_1( \bar{s}^1)} q_i + \sum_{i\in I_2( \bar{s}^1)} \frac{\lambda \bar{s}^1}{2\alpha_i -\lambda/m} + \sum_{i\in I_3( \bar{s}^1)} Q_i , \label{eq_s_1}\\
    m \bar{s}^2 & = \sum_{i\in I_1( \bar{s}^2)} q_i + \sum_{i\in I_2( \bar{s}^2)} \frac{\lambda \bar{s}^2}{2\alpha_i -\lambda/m} + \sum_{i\in I_3( \bar{s}^2)} Q_i. \label{eq_s_2}
\end{align}
Since $\bar{s}^1 < \bar{s}^2$, \eqref{eq_inclusion} yields $ I_1(\bar{s}^2) \subseteq I_1(\bar{s}^1)$ and $I_3(\bar{s}^1) \subseteq I_3(\bar{s}^2)$. Combining this with \eqref{eq_union}, we obtain
\begin{equation}\label{eq_relation}
    I_2(\bar{s}^1) \cup I_1(\bar{s}^1) \setminus I_1(\bar{s}^2) =  I_2(\bar{s}^2) \cup I_3(\bar{s}^2) \setminus I_3(\bar{s}^1) = \colon I.
\end{equation}
By \eqref{eq_beta_i-1} and \eqref{eq_beta_i-3}, it holds:
\begin{align}
    q_i &\geqslant \frac{\lambda \bar{s}^1}{2\alpha_i - \lambda/m}, \quad \text{for } i \in I_1(\bar{s}^1) \setminus I_1(\bar{s}^2), \label{eq_q} \\
    Q_i &\leqslant \frac{\lambda \bar{s}^2}{2\alpha_i - \lambda/m}, \quad \text{for } i \in I_3(\bar{s}^2) \setminus I_3(\bar{s}^1). \label{eq_Q}
\end{align}
Substituting \eqref{eq_q} into \eqref{eq_s_1} and \eqref{eq_Q} into \eqref{eq_s_2}, and using \eqref{eq_relation}, we obtain
\begin{align}
    m \bar{s}^1 & \geqslant \sum_{i\in I_1( \bar{s}^2)} q_i + \sum_{i\in I} \frac{\lambda \bar{s}^1}{2\alpha_i -\lambda/m} + \sum_{i\in I_3( \bar{s}^1)} Q_i , \label{eq_ms_1}\\
    m \bar{s}^2 & \leqslant \sum_{i\in I_1( \bar{s}^2)} q_i + \sum_{i\in I} \frac{\lambda \bar{s}^2}{2\alpha_i -\lambda/m} + \sum_{i\in I_3( \bar{s}^1)} Q_i. \label{eq_ms_2}
\end{align}
Taking the difference between \eqref{eq_ms_2} and \eqref{eq_ms_1}, and noting that $\bar{s}^1<\bar{s}^2$, we have
\begin{equation}\label{eq_contradiction_2}
    1 \leqslant \sum_{i\in I} \frac{\lambda}{2m\alpha_i - \lambda}. 
\end{equation}
On the other hand, \eqref{eq_ms_1} implies that if $I_1(\bar{s}^2) \cup I_3( \bar{s}^1) \neq \emptyset$, then 
\begin{equation*}
    1 > \sum_{i\in I} \frac{\lambda}{2m\alpha_i - \lambda}.
\end{equation*}
This contradicts \eqref{eq_contradiction_2}. Therefore, $I_1(\bar{s}^2) \cup I_3( \bar{s}^1) = \emptyset$, so $I = [m]$. Consequently, \eqref{eq_ms_1} and \eqref{eq_contradiction_2} give
\begin{equation*}
    1 = \sum_{i\in [m]} \frac{\lambda}{2m\alpha_i - \lambda},
\end{equation*}
which contradicts the assumption $\lambda \neq \lambda^*$, by the definition of $\lambda^*$ in \eqref{eq_lambda_star}. Hence the NE is unique in this case.

\noindent\textbf{Step 2} (The case $\lambda\geqslant \bar{\lambda}$). 
Assume that $s^1$ and $s^2$ are two distinct NEs and that $\bar{s}^1 < \bar{s}^2$.
Arguing as in \eqref{eq_s_1}-\eqref{eq_s_2} and using $\lambda \geqslant \bar{\lambda}$, we obtain
\begin{align*}
    m \bar{s}^1 & = \sum_{i\in I_1( \bar{s}^1)} q_i + \sum_{i\in I_2( \bar{s}^1)} \frac{\lambda \bar{s}^1}{2\alpha_i -\lambda/m} + \sum_{i\in I_3( \bar{s}^1)} Q_i + \sum_{i\in I_4( \bar{s}^1)} Q_i , \\
    m \bar{s}^2 & = \sum_{i\in I_1( \bar{s}^2)} q_i + \sum_{i\in I_2( \bar{s}^2)} \frac{\lambda \bar{s}^2}{2\alpha_i -\lambda/m} + \sum_{i\in I_3( \bar{s}^2)} Q_i + \sum_{i\in I_4( \bar{s}^2)} Q_i. 
\end{align*}
Since $I_4(x)$ is independent of $x$, we have $I_4(\bar{s}^1) = I_4(\bar{s}^2) \neq \emptyset$. By the same argument as in Step 1, we obtain \eqref{eq_contradiction_2}, namely,
\begin{equation*}
    1 \leqslant \sum_{i\in I} \frac{\lambda}{2m\alpha_i - \lambda}. 
\end{equation*}
In addition, an argument analogous to \eqref{eq_ms_1} gives
\begin{align*}
    m \bar{s}^1  & \geqslant \sum_{i\in I_1( \bar{s}^2)} q_i + \sum_{i\in I} \frac{\lambda \bar{s}^1}{2\alpha_i -\lambda/m} + \sum_{i\in I_3( \bar{s}^1)} Q_i + \sum_{i\in I_4( \bar{s}^1)} Q_i\\
    & \geqslant \sum_{i\in I} \frac{\lambda \bar{s}^1}{2\alpha_i -\lambda/m}  + \sum_{i\in I_4( \bar{s}^1)} Q_i \\
    & > \sum_{i\in I} \frac{\lambda \bar{s}^1}{2\alpha_i -\lambda/m},
\end{align*}
where the last inequality is strict because $I_4(\bar{s}^1)$ is nonempty. This contradicts \eqref{eq_contradiction_2}, and therefore the NE is unique in this case.

\noindent\textbf{Step 3} (The case $\lambda = \lambda^*$ and $c_1 \geqslant c_2$). Assume that $s^1$ and $s^2$ are two distinct NEs. By the argument in Step 1, we obtain $I_1(\bar{s}^2) \cup I_3( \bar{s}^1) = \emptyset$. Using $I_3( \bar{s}^1) = \emptyset$ in \eqref{eq_s_1}, and recalling that $\lambda^*$ solves \eqref{eq_lambda_star}, we obtain
\begin{equation*}
    \sum_{i\in I_1(\bar{s}^1)}  \frac{\lambda^* \bar{s}^1}{2\alpha_i - \lambda^*/m} = \sum_{i\in I_1(\bar{s}^1)}  q_i.
\end{equation*}
Combining with the definition of $I_1(\bar{s}^1)$, it follows that
\begin{equation*}
    \frac{\lambda^*\bar{s}^1}{2\alpha_i - \lambda^*/m} =  q_i, \quad \text{for } i\in I_1(\bar{s}^1).
\end{equation*}
Given the fact that $I_3(\bar{s}^1) = \emptyset$, we deduce that
\begin{equation*}
    q_i\leqslant  \frac{\lambda^* \bar{s}^1 }{2\alpha_i - \lambda^*/m} < Q_i, \quad \text{for } i\in [m].
\end{equation*}
This is equivalent to 
\begin{equation*}
    \max_{i\in [m]}\left\{ \frac{2\alpha_iq_i}{\lambda^*} -\frac{q_i}{m} \right\} \leqslant \bar{s}^1 , \quad \text{and} \quad \bar{s}^1 < \min_{i\in [m]}\left\{\frac{2\alpha_iQ_i}{\lambda^*} -\frac{Q_i}{m}\right\}.
\end{equation*}
Recalling the definitions of $c_1$ and $c_2$ in \eqref{constant_C1_C2}, the previous inequality yields $c_1\leq \bar{s}^1 < c_2$, which contradicts the assumption $c_1\geqslant c_2$. Therefore, the NE is unique.

\noindent\textbf{Step 4} (The case $\lambda = \lambda^*$ and $c_1 < c_2$). Take any $s^*\in \hat{\mathcal{S}}$ as in \eqref{eq_infinit_NE}. Thus, for some $c\in (c_1,c_2)$,
\begin{equation*}
    s^{*}_i = \frac{\lambda^* c}{2 \alpha_i-\lambda^* /m} \,, \quad  \forall i \in [m].
\end{equation*}
It follows that
\begin{equation*}
    \bar{s}^{*} = \frac{1}{m}\sum_{i\in [m]} s^*_i = c \in (c_1,c_2).
\end{equation*}
By direct calculation,
\begin{equation*}
    \frac{\lambda^*\left(\bar{s}^*-s^*_i/m\right)}{2\left(\alpha_i-\lambda^*/m\right)} =  \frac{c \lambda^*}{2\alpha_i - \lambda^*/m}, \quad \text{for } i\in [m].
\end{equation*} 
Since $c\in (c_1,c_2)$, it follows that
\begin{equation*}
    q_i < \frac{\lambda^*\left(\bar{s}^*-s^*_i/m\right)}{2\left(\alpha_i-\lambda^*/m\right)} < Q_i, \quad \text{for } i\in [m].
\end{equation*}
Combining this with \eqref{eq_lambda_star}, we deduce that $s^{*}$ satisfies the fixed-point system \eqref{eq_s_i-1}-\eqref{eq_s_i-3}, and hence $s^{*}$ is an NE\@. This proves the sufficiency of \eqref{eq_infinit_NE}.

Since $c$ can be chosen arbitrarily in $(c_1,c_2)$, the game $\hat{\Gamma}_{\textnormal{FL}}$ has infinitely many NEs.
In particular, the game $\hat{\Gamma}_{\textnormal{FL}}$ has at least two distinct NEs. Then, by an argument similar to that in Step 3, we deduce that for any NE of $\hat{\Gamma}_{\textnormal{FL}}$, denoted by $s'$, it holds that
\begin{equation}\label{eq_s'}
    q_i\leqslant  \frac{\lambda^* \bar{s}' }{2\alpha_i - \lambda^*/m} \leqslant  Q_i, \quad \text{for } i\in [m],
\end{equation}
where $\bar{s}' = \sum_{i\in [m]} s'_i/m$. It also follows from \eqref{eq_s'} and \eqref{constant_C1_C2} that $\bar{s}' \in (c_1,c_2)$.
Moreover, by \eqref{eq_best_response}, $s'_i = \beta_i(\bar{s}')$ for all $i$. Combining with \eqref{eq_s'}, we obtain that
\begin{equation*}
    s_i' =  \frac{\lambda^* \bar{s}' }{2\alpha_i - \lambda^*/m} \quad \text{for } i\in [m].
\end{equation*}
The necessity of condition \eqref{eq_infinit_NE} follows.
\end{proof}

\begin{proof}[Proof of Corollary~\ref{cor_uniq}.]
We combine Theorem~\ref{thm_uniq} with the fixed-point representation established above.
For small and large values of $\lambda$, the boundary profiles $(q_i)_{i=1}^m$ and $(Q_i)_{i=1}^m$
satisfy the equilibrium conditions directly.
For the intermediate regimes, the same partition argument yields the bounds
$\bar{s}^*\le c_1$ for $\lambda<\lambda^*$ and $\bar{s}^*\ge c_2$ for $\lambda>\lambda^*$.

\noindent\textbf{Step 1} (Proof of points (1) and (4)).
Assume that $\lambda\in (0,\lambda_1)$. Since $\lambda_1\leq \lambda^*$, Theorem~\ref{thm_uniq} implies that the NE is unique. It therefore suffices to prove that $(q_i)_{i=1}^m$ is an NE\@. Let $\bar{q} = \sum_{i\in [m]} q_i/m$. By the definition of $\lambda_1$ in \eqref{eq_lambda_1_2}, we obtain
\begin{equation}
    q_i \geqslant \frac{\lambda (\bar{q} - q_i/m)}{2(\alpha_i-\lambda/m)},\quad \text{for } i\in [m].
\end{equation}
Therefore, $(q_i)_{i=1}^m$ is an NE because it satisfies the fixed-point system \eqref{eq_s_i-1}-\eqref{eq_s_i-3}. The case $\lambda \in (\lambda_2, +\infty)$ is proved in the same way, using the definition of $\lambda_2$ in \eqref{eq_lambda_1_2}.

\noindent \textbf{Step 2} (Proof of point (2)).
Assume that $\lambda < \lambda^{*}$. Let $s^*$ be the unique NE, and $\bar{s}^* = \sum_{i\in [m]} s^*_i/m$. Since $\lambda^* < \bar{\lambda}$, we have $I_4(\bar{s}^*) = \emptyset$.
By \eqref{eq_fixed_point}, we obtain that
\begin{equation*}
    m \bar{s}^* = \sum_{i\in I_1( \bar{s}^*)} q_i + \sum_{i\in I_2( \bar{s}^*)} \frac{\lambda \bar{s}^*}{2\alpha_i -\lambda/m} + \sum_{i\in I_3( \bar{s}^*)} Q_i.
\end{equation*}
If $I_1(\bar{s}^*) = \emptyset$, then
\begin{equation*}
    m \bar{s}^* =  \sum_{i\in I_2( \bar{s}^*) 
    } \frac{\lambda \bar{s}^*}{2\alpha_i -\lambda/m} + \sum_{i\in I_3( \bar{s}^*)} Q_i \leqslant \sum_{i\in [m]
    } \frac{\lambda \bar{s}^*}{2\alpha_i -\lambda/m}.
\end{equation*}
This implies
\begin{equation*}
    1 \leq \sum_{i\in [m]
    } \frac{\lambda }{2\alpha_i -\lambda/m},
\end{equation*}
which contradicts the assumption $\lambda<\lambda^*$. Therefore, $I_1(\bar{s}^*) \neq \emptyset$. Hence
\begin{equation*}
     m \bar{s}^* \leqslant \sum_{i\in I_1( \bar{s}^*)} q_i + \sum_{i\in [m]\setminus  I_1( \bar{s}^*)}\frac{\lambda \bar{s}^*}{2\alpha_i -\lambda/m} \leqslant \sum_{i\in I_1( \bar{s}^*)} q_i + \sum_{i\in [m]\setminus  I_1( \bar{s}^*)}\frac{\lambda^* \bar{s}^*}{2\alpha_i -\lambda^*/m}.
\end{equation*}
Suppose that 
\begin{equation}\label{eq_s_star_c1}
    \bar{s}^{*} > c_1 = \max_{i \in[m]}\left\{\frac{2 \alpha_i q_i}{\lambda^*}-\frac{q_i}{m}\right\}.
\end{equation}
Then, 
\begin{equation*}
     1 < \sum_{i\in I_1( \bar{s}^*)} \frac{q_i}{m \left(\frac{2 \alpha_i q_i}{\lambda^*}-\frac{q_i}{m}\right)} + \sum_{i\in [m]\setminus  I_1( \bar{s}^*)}\frac{\lambda^*}{2m \alpha_i -\lambda^*}=\sum_{i\in [m]} \frac{\lambda^* }{2 m \alpha_i -\lambda^* } =1,
\end{equation*}
where the first inequality is strict because $I_1(\bar{s}^*)$ is nonempty.
This is a contradiction. Therefore, \eqref{eq_s_star_c1} cannot hold, and thus $\bar{s}^* \leqslant c_1$.

\noindent\textbf{Step 3} (Proof of point (3)). Assume that $\lambda > \lambda^{*}$. Let $s^*$ be the unique NE, and let $\bar{s}^* = \sum_{i\in [m]} s^*_i/m$. Then
\begin{equation*}
    m \bar{s}^* = \sum_{i\in I_1( \bar{s}^*)} q_i + \sum_{i\in I_2( \bar{s}^*)} \frac{\lambda \bar{s}^*}{2\alpha_i -\lambda/m} + \sum_{i\in I_3( \bar{s}^*)} Q_i + \sum_{i\in I_4( \bar{s}^*)} Q_i.
\end{equation*}
If $ I_4(\bar{s}^*) = \emptyset$, then an argument similar to that in Step 2 shows that $I_3(\bar{s}^*) \neq \emptyset$. Repeating the proof of Step 2, we deduce that $\bar{s}^* \geqslant c_2$.

Now assume that $I_3(\bar{s}^*) = \emptyset$ and $I_4(\bar{s}^*)\neq \emptyset$. Then
\begin{equation*}
    m \bar{s}^* \geqslant \sum_{i\in I_1(\bar{s}^*) \cup I_2( \bar{s}^*)} \frac{\lambda \bar{s}^*}{2\alpha_i -\lambda/m} + \sum_{i\in I_4( \bar{s}^*)} Q_i =   \sum_{i\in [m]\setminus I_4(\bar{s}^*)} \frac{\lambda \bar{s}^*}{2\alpha_i -\lambda/m} + \sum_{i\in I_4( \bar{s}^*)} Q_i.
\end{equation*}
Since $2\alpha_i - \lambda/m > 0$ for $i \notin I_4(\bar{s}^*)$, the monotonicity of $\lambda / (2m \alpha_i - \lambda)$ yields
\begin{equation}\label{eq_m_s_star}
     1 \geqslant    \sum_{i\in [m]\setminus I_4(\bar{s}^*)} \frac{\lambda^* }{2 m \alpha_i -\lambda^* } + \sum_{i\in I_4( \bar{s}^*)} \frac{Q_i}{m \bar{s}^*}.
\end{equation}
Suppose that 
\begin{equation}\label{eq_s_star_c2}
    \bar{s}^{*} < c_2 = \min_{i \in[m]}\left\{\frac{2 \alpha_i Q_i}{\lambda^*}-\frac{Q_i}{m}\right\}.
\end{equation}
Substituting \eqref{eq_s_star_c2} into \eqref{eq_m_s_star}, we obtain that
\begin{equation*}
    1 > \sum_{i\in [m]\setminus I_4(\bar{s}^*)} \frac{\lambda^* }{2 m \alpha_i -\lambda^* } + \sum_{i\in I_4( \bar{s}^*)} \frac{Q_i}{m \left(\frac{2 \alpha_i Q_i}{\lambda^*}-\frac{Q_i}{m}\right)} = \sum_{i\in [m]} \frac{\lambda^* }{2 m \alpha_i -\lambda^* } =1.
\end{equation*}
This is a contradiction. Therefore, \eqref{eq_s_star_c2} cannot hold, and thus $\bar{s}^* \geqslant c_2$.
\end{proof}

\subsection{Convergence of the Best-Response Algorithm for Potential Games}\label{subsec_proof_part2}
This subsection establishes a general convergence result for Algorithm~\ref{alg_BRA} when it is applied to a weighted potential game $\Gamma$, with the obvious adaptations of the algorithm to this setting. The argument is independent of the FL context discussed elsewhere in the paper. All notation introduced here is local to this subsection.

Consider a weighted potential game $\Gamma$ as in Definition~\ref{def_potential_game}, with strategy sets $\mathcal{S}_i \subseteq \mathbb{R}^d$, payoff functions $P_i(s_i, s_{-i})$, and a $w$-potential $P$, where $w_i > 0$ for all $i \in [m]$. To prove the convergence of Algorithm~\ref{alg_BRA}, we impose Assumption~\ref{ass_continuous} on the convexity and regularity of $\mathcal{S}_i$ and $P_i$. The proof of Theorem~\ref{thm_BRA_general} is inspired by convergence arguments for block coordinate descent in multi-convex optimization and for congestion games, see \citep{xu2013block} and \citep[Sec.\@ 3.2]{liu2023approximate}.

\begin{assumption}\label{ass_continuous}
The following conditions hold.
\begin{enumerate}
    \item The strategy sets $\mathcal{S}_i$ are compact and convex subsets of $\mathbb{R}^d$.
    \item For each $i \in [m]$, there exists a constant $\alpha_i > 0$ such that $P_i(s_i, s_{-i})$ is $\alpha_i$-strongly concave with respect to $s_i$ for every $s_{-i} \in \mathcal{S}_{-i}$. We also define $\alpha_{\text{min}} = \min_{i \in [m]} \{\alpha_i\}$.
    \item For any $i \in [m]$, the payoff function $P_i(s_i, s_{-i})$ is continuously differentiable with respect to $s_i$, and this partial gradient is denoted by $\nabla_i P_i(s_i, s_{-i})$.
    \item There exists a constant $L > 0$ such that for any $i \in [m]$, $\nabla_i P_i(s_i, s_{-i})$ is $L$-Lipschitz with respect to $(s_i, s_{-i})$.
\end{enumerate}
\end{assumption}

Before stating the convergence result, we recall the definition of an $\epsilon$-NE.
\begin{definition}[$\epsilon$-NE]\label{def:eNE} 
In a game $\Gamma$,
for any $\epsilon\geqslant 0$, we call a point $s^{\epsilon}\in \mathcal{S}$ an $\epsilon$-NE if the following inequality holds:
\begin{equation*}
    P_i\left(s_i^{\epsilon}, s_{-i}^{\epsilon}\right) \geqslant P_i\left(s_i, s_{-i}^\epsilon\right) -\epsilon, \quad  \forall s_i \in \mathcal{S}_i, \, \forall i \in [m].
\end{equation*}
In particular, if $\epsilon=0$, then $s^{0}$ is an NE.
\end{definition}

\begin{theorem}\label{thm_BRA_general}
Let $\Gamma$ be a $w$-potential game, and suppose that Assumption~\ref{ass_continuous} holds. Let $ s^k$ be the output of Algorithm~\ref{alg_BRA} at iteration $k$, for any $k\geqslant 1$. Then the following statements hold.
\begin{enumerate}
    \item  
    For any $K\geqslant 1$, there exists $k\in\{1,\ldots,K\}$ such that $s^k$ is an $\mathcal{O}(1/K)$-NE.
    \item Any cluster point of the sequence $\{s^k\}_{k\geqslant 1}$ is an NE of $\Gamma$.
\end{enumerate}
\end{theorem}
\begin{proof}
The proof follows the standard block coordinate ascent viewpoint on the potential.
Each best-response update increases the potential by at least a quadratic quantity, which implies
that $\sum_k\|s^k-s^{k-1}\|^2$ is finite.
Consequently, there exists an iterate among the first $K$ steps whose update gap is small enough to imply an $\epsilon$-NE estimate with $\epsilon=\mathcal{O}(1/K)$.
Finally, along any convergent subsequence, the update gap vanishes, so the limiting profile satisfies
the exact best-response inequalities and is therefore an NE.

\noindent\textbf{Step 1} (Finite sum of squares). Let us define $p_i^k\colon \mathcal{S}_i\to \mathbb{R}$,
\begin{equation*}
    p_i^k(s_i) = P_{i}(s_1^k,\ldots, s_{i-1}^k, s_i, s_{i+1}^{k-1}, \ldots, s^{k-1}_m).
\end{equation*}
Let $P$ be a $w$-potential of $\Gamma$.  Then, for any $i$,
\begin{align*}
    \frac{1}{w_i} \left(p_i^k(s_i^k) - p_i^k(s_i^{k-1}) \right)  =\, & P(s_1^k,\ldots, s_{i-1}^k, s_i^k, s_{i+1}^{k-1}, \ldots, s^{k-1}_m) \\
    & - P(s_1^k,\ldots, s_{i-1}^k, s_i^{k-1}, s_{i+1}^{k-1}, \ldots, s^{k-1}_m).
\end{align*}
Summing the previous equality over $i$, we obtain that
\begin{equation*}
    \sum_{i\in[m]}  \frac{1}{w_i} \left(p_i^k(s_i^k) - p_i^k(s_i^{k-1}) \right)= P(s^k) - P(s^{k-1}).
\end{equation*}
On the other hand, by the strong concavity of $P_i$ with respect to $s_i$ and the fact that $s^k_i$ is the maximizer of $p_i^k$, we deduce that
\begin{equation*}
    p_i^k(s_i^k) - p_i^k(s_i^{k-1}) \geqslant \frac{\alpha_i}{2} \|s_i^k - s_i^{k-1}\|^2.
\end{equation*}
It follows that
\begin{equation*}
    \sum_{i=1}^m\frac{\alpha_i}{2 w_i} \|s_i^k - s_i^{k-1}\|^2 \leqslant   P(s^k) - P(s^{k-1}).
\end{equation*}
Summing the previous inequality over $k$ from $1$ to $K$, we deduce that
\begin{equation*}
    \sum_{k=1}^K \|s^k-s^{k-1}\|^2 \leqslant \frac{2}{\delta} \left(   P(s^{K}) - P(s^{0}) \right)\leqslant \frac{2}{\delta} \left(   P^{*} - P(s^{0}) \right),
\end{equation*}
where $\delta = \min_{i \in [m]} \{\alpha_i/w_i\} > 0$, and $P^{*}$ is the maximum value of $P$ and is finite.

\noindent\textbf{Step 2} (Proof of point (1)). Let us assume that $ \|s^{k} - s^{k-1}\| \leqslant \epsilon $ for some $k\geqslant 1$ and $\epsilon >0$. 
Then, by the strong concavity of $P_i$ with respect to $s_i$, we have
\begin{equation*}
    P_i(s_i, s_{-i}^{k}) - P_i(s_i^{k}, s_{-i}^{k}) \leqslant  \langle \nabla_i P_i (s_{i}^{k}, s_{-i}^{k}), s_i - s_{i}^{k} \rangle - \frac{\alpha_{\text{min}}}{2}\|s_i-s_{i}^k\|^2.
\end{equation*}
We decompose the linear term on the right-hand side of the previous inequality in the following way:
\begin{equation*}
    \langle \nabla_i P_i (s_{i}^{k}, s_{-i}^{k}), s_i - s_{i}^{k} \rangle = \gamma_1 + \gamma_2,
\end{equation*}
where 
\begin{align*}
    \gamma_1 &= \langle \nabla_i P_i (s_{i}^{k}, s_{-i}^{k}) -  \nabla p_i^k(s_i^{k}), s_i - s_{i}^{k} \rangle,\\
    \gamma_2 &= \langle   \nabla p_i^k(s_i^{k}), s_i - s_{i}^{k} \rangle.
\end{align*}
Since $s_i^k$ is the maximizer of $p_i^k$, the first-order optimality condition implies that $\gamma_2\leqslant 0$. Moreover, by the definition of $p_i^k$, we have
\begin{equation*}
    \nabla p_i^k(s_i^{k}) = \nabla_i P_i(s_1^{k},\ldots, s_{i-1}^k, s_i^k, s_{i+1}^{k-1},\ldots, s^{k-1}_m).
\end{equation*}
By the Lipschitz continuity of $\nabla_i P_i$, we deduce that
\begin{equation*}
    \|\nabla_i P_i (s_{i}^{k}, s_{-i}^{k}) -  \nabla p_i^k(s_i^{k})\| \leqslant L \|s^k -(s_1^{k},\ldots, s_{i-1}^k, s_i^k, s_{i+1}^{k-1},\ldots, s^{k-1}_m) \| \leqslant L \|s^k - s^{k-1}\|\leqslant L\epsilon.
\end{equation*}
Therefore, for any $i$ and $s_i\in \mathcal{S}_i$, 
\begin{equation*}
    P_i(s_i, s_{-i}^{k}) - P_i(s_i^{k}, s_{-i}^{k}) \leqslant L\epsilon \|s_i-s_i^k\| - \frac{\alpha_{\text{min}}}{2}\|s_i-s_i^{k}\|^2 \leqslant \frac{L^2 \epsilon^2}{2\alpha_{\text{min}}}.
\end{equation*}
Therefore, we prove that $s^k$ is an $(L^2 \epsilon^2/(2\alpha_{\text{min}}))$-NE\@. From step 1, fixing any $K\geqslant 1$, there exists $k_0\leqslant K$ such that
\begin{equation*}
    \|s^{k_0}-s^{k_0-1}\|^2 \leqslant \frac{2}{\delta K} (P^{*}-P(s^0)),
\end{equation*}
where $P^{*}$ is the maximum value of $P$ over $\mathcal{S}$ and is finite.
Taking $k=k_0$, it follows that $s^{k_0}$ is an $\epsilon'$-NE, where
\begin{equation*}
    \epsilon' =  \frac{L^2}{2\alpha_{\text{min}}} \frac{2}{\delta K} (P^{*}-P(s^0)) = \frac{L^2 (P^{*}-P(s^0) )}{ \delta \, \alpha_{\text{min}} } \frac{1}{K}.
\end{equation*}

\noindent \textbf{Step 3} (Proof of point (2)).
By the compactness of $\mathcal{S}$, assume that a subsequence of $\{s^{k}\}_{k\geqslant 1}$, denoted by $\{s^{\varphi(k)}\}_{k\geqslant 1}$, converges to some point $s^{*}\in \mathcal{S}$. By step 1, we have that $\|s^{\varphi(k)-1} - s^{\varphi(k)}\|\to 0$. Similar to step 2, we have
\begin{equation*}
    P_i(s_i, s_{-i}^{*}) - P_i(s_i^{*}, s_{-i}^{*}) \leqslant  \langle \nabla_i P_i (s_{i}^{*}, s_{-i}^{*}), s_i - s_{i}^{*} \rangle - \frac{\alpha_{\text{min}}}{2}\|s_i-s_{i}^*\|^2.
\end{equation*}
The previous first-order term is decomposed by
\begin{align*}
    \langle \nabla_i P_i (s_{i}^{*}, s_{-i}^{*}), s_i - s_{i}^{*} \rangle 
    = \, & \langle \nabla_i P_i (s_{i}^{\varphi(k)}, s_{-i}^{\varphi(k)}), s_i - s_{i}^{\varphi(k)} \rangle +  \langle \nabla_i P_i (s_{i}^{*}, s_{-i}^{*}), s_i^{\varphi(k)} - s_{i}^{*} \rangle\\
    & + \langle \nabla_i P_i (s_{i}^{*}, s_{-i}^{*}) - \nabla_i P_i (s_{i}^{\varphi(k)}, s_{-i}^{\varphi(k)}) , s_i-s_i^{\varphi(k)}  \rangle \\
    \leqslant \,  & L \left(\|s^{\varphi(k)}- s^{\varphi(k)-1}\| + \|s^{*}-s^{\varphi(k)}\| \right) \|s_i-s_{i}^{\varphi(k)}\|
    \\
    & + \| \nabla_i P_i (s^{*})\| \| s_i^{\varphi(k)} - s_{i}^{*} \|,
\end{align*}
where the last inequality follows from a similar argument to step 2 and the Lipschitz continuity of $\nabla_iP_i$. By passing $k$ to infinity, we deduce that $P_i(s_i, s_{-i}^{*}) - P_i(s_i^{*}, s_{-i}^{*}) \leqslant 0$ for any $i$ and $s_i$. Therefore, $s^{*}$ is an NE.
\end{proof}

\section{Conclusions and Future Work}\label{sec_conclusions}

In this paper, we introduce a potential game framework to model the interplay among training costs, incentives, and effort selection in federated learning (FL). We prove the existence of Nash equilibria (NEs) in the proposed game and further investigate the uniqueness of the NE in a stationary scenario where clients maintain constant efforts throughout training. Additionally, we characterize how the NEs evolve with the server's reward factor, identifying three critical thresholds: the activation point, the jump point, and the saturation point. Notably, our findings highlight the significance of the jump point, at which clients' training efforts increase sharply, suggesting that it provides a natural target for server-side reward tuning. Extensive numerical experiments further support our theoretical results.
Overall, our work provides a robust foundation for understanding the interplay between clients' rational training efforts and the server's incentives.

\medskip

For future work, we propose the following perspectives.
\begin{enumerate}
\item A first extension of our game model could focus on incorporating a discount factor into the server's reward scheme based on the number of rounds. This aligns with the convergence properties of widely used stochastic gradient algorithms in machine learning, where significant gains in model performance are typically observed during the early training stages \citep[Sec.\@ 4]{bottou2018optimization}. This phenomenon is also reflected in our simulation results shown in \Cref{fig_FL_train}, where major improvements occur in the first several rounds. Furthermore, adding a discount factor may also help address the issue of non-uniqueness of the NE in heterogeneous cases, as the counterexample mentioned in Remark~\ref{rem_non_unique} no longer applies in this scenario.

\item A second direction focuses on the asymptotic behavior of our FL game as the number of clients $m$ approaches infinity. In this scenario, a nonatomic version of $\Gamma_{\text{FL}}$ or $\hat{\Gamma}_{\text{FL}}$ could be analyzed. The NEs in this setting are characterized by an optimality condition over the distribution of training efforts among a continuum of clients, see \citep[Thm.\@ 1]{cheung2018nonatomic}. It is worth noting that proving the uniqueness of the NE in the nonatomic case is generally challenging when the price increases with respect to the aggregate, as discussed in \citep[Prop.\@ 3]{cheung2018nonatomic}.
The techniques used to prove the uniqueness results in Section~\ref{sec_proof} can be naturally adapted to these nonatomic games.

\item Extending our game model to other aspects of FL, such as fairness \citep{donahue2021optimality} and privacy \citep{zhang2024game}, is a promising avenue for future research. 
For example, in a differential privacy scenario, each client's strategy can be the standard deviation of their noise. Therefore, we can model the trade-off between data privacy (which increases with the noise level) and model accuracy (which decreases with the noise level) using our FL game, with the necessary adaptations.

\item {Recent studies identify open challenges in federated foundation models \citep{fan2025ten} and federated reasoning-oriented large language models (LLMs) \citep{wei2025federated}. Our server-tunable potential game framework naturally extends to these settings by modeling clients' adaptation and reasoning budgets as multidimensional efforts and shaping rewards through global utility proxies such as validation gains, calibration or consistency scores, and safety metrics under bandwidth and risk constraints. Central research questions include the existence and uniqueness of equilibria with coupled efforts and the design of optimal incentives under active alignment or safety constraints.}

\item Finally, beyond FL, extending our game model to other applications is also of interest. The current game formulation should be well suited to modeling systems with positive consumption externalities \citep{katz1985network}. For example, in technology and social media platforms, the value of the platform to each user increases as more users participate. As a result, such platforms often design incentives that scale with total engagement, thereby reinforcing collective participation.

\end{enumerate}

\section*{Acknowledgments}
The authors' names are listed in alphabetical order by family name to signify equal contributions.
The work was partially supported by the European Research Council (ERC) under the European Union's Horizon 2030 research and innovation programme (grant agreement NO: 101096251-CoDeFeL);  by the Alexander von Humboldt Professorship program; the European Union's Horizon Europe MSCA project Modelling and control of flexible structures interacting with fluids (ModConFlex; grant agreement No. 101073558; call HORIZON-MSCA-2021-DN-01-01; project DOI: \href{https://doi.org/10.3030/101073558}{10.3030/101073558}); the Transregio 154 Project "Mathematical Modelling, Simulation and Optimization Using the Example of Gas Networks" of the DFG; the AFOSR 24IOE027 project; the SURE-AI Norwegian Centre for Sustainable, Risk-Averse, and Ethical AI grant 357482, Research Council of Norway;  by the Grant PID2023-146872OB-I00-DyCMaMod of MICIU (Spain) and by the COST Actions CA24122 - Multiscale Stochastics, Patterns, and Analysis of Combinatorial Environments and  CA24136 - Interactions between Control Theory and Machine Learning.

\bibliographystyle{elsarticle-harv} 
\bibliography{references.bib}

\end{document}